\documentclass{article}
\usepackage{graphicx}
\usepackage{amsmath} 
\usepackage[sglblindworkshop]{neurips_2025}
\usepackage[utf8]{inputenc} % allow utf-8 input
\usepackage[T1]{fontenc}    % use 8-bit T1 fonts
\usepackage{hyperref}       % hyperlinks
\usepackage{url}            % simple URL typesetting
\usepackage{booktabs}       % professional-quality tables
\usepackage{amsfonts}       % blackboard math symbols
\usepackage{nicefrac}       % compact symbols for 1/2, etc.
\usepackage{microtype}      % microtypography
\usepackage{xcolor}         % colors

\title{GRAIL: A Benchmark for \underline{GR}aph \underline{A}ct\underline{I}ve  \underline{L}earning in Dynamic Sensing Environments}

\author{
  Maryam Khalid \\
  Amazon\\
  Seattle, WA, USA.\\
  \texttt{maryamkd@amazon.com} \\
  \And
  Akane Sano \\
  Department of Electrical Computer Engineering \\
  Rice University \\
  Houston, TX, USA.
\texttt{akane.sano@rice.edu} \\
}

\begin{document}

\maketitle
\nolinenumbers

\begin{abstract}

Graph-based Active Learning (AL) leverages the structure of graphs to efficiently prioritize label queries, reducing labeling costs and user burden in applications like health monitoring, human behavior analysis, and sensor networks. By identifying strategically positioned nodes, graph AL minimizes data collection demands while maintaining model performance, making it a valuable tool for dynamic environments. Despite its potential, existing graph AL methods are often evaluated on static graph datasets and primarily focus on prediction accuracy, neglecting user-centric considerations such as sampling diversity, query fairness, and adaptability to dynamic settings. To bridge this gap, we introduce GRAIL, a novel benchmarking framework designed to evaluate graph AL strategies in dynamic, real-world environments. GRAIL introduces novel metrics to assess sustained effectiveness, diversity, and user burden, enabling a comprehensive evaluation of AL methods under varying conditions. Extensive experiments on datasets featuring dynamic, real-life human sensor data reveal trade-offs between prediction performance and user burden, highlighting limitations in existing AL strategies. GRAIL demonstrates the importance of balancing node importance, query diversity, and network topology, providing an evaluation mechanism for graph AL solutions in dynamic environments.
\end{abstract}

\section{Introduction}

In the realm of health and well-being monitoring, mobile and wearable devices continuously collect data to track user behaviors, physiological metrics, and contextual information, often paired with Ecological Momentary Assessments (EMAs) or periodic surveys that generates ground truth labels essential for model training and validation. However, frequent prompts for EMAs can impose a substantial burden on users, potentially leading to disengagement or incomplete data due to survey fatigue. %Reducing this user burden while maintaining high-quality labeled data is a pivotal challenge for applications reliant on dynamic sensor data. 
A similar challenge applies to dynamic networked environments, such as social networks, sensor networks, or Internet of Things (IoT) applications, where the continuous operation and data demands on individual nodes (e.g., users, sensors, or devices) can lead to resource depletion, sensor battery drain, or data overload, impacting overall system performance and reliability.

Graph-based AL offers a promising approach to address this challenge by leveraging the structure of user networks to selectively query only a subset of users or sensors for data collection.  This approach typically integrates two main components: AL strategies and graph-embedding techniques, balancing performance metrics through an optimization function such as:

\[
x^* = \arg\min_x \mathcal{L}(x)
\]
\[
\mathcal{L} = f(U, C, N)
\]

where \( x \) is the sample to query and \( \mathcal{L} \) is a loss function based on a combination of metrics: uncertainty \( U \), clustering \( C \), and network metrics \( N \). Uncertainty is often measured using entropy, mutual information, or prediction confidence, guiding the model to query points with the highest ambiguity \cite{AL25, AL26, AL29}. Clustering metrics leverage the graph structure to identify representative nodes, using techniques such as clustering coefficient, K-Medoids clustering, and modularity clustering \cite{AL25, AL27, AL28, AL30}. Network metrics like degree centrality and pagerank centrality prioritize influential nodes, optimizing model performance even with limited labeled data \cite{AL25, AL26}. 

Despite their contributions, existing evaluations of these approaches exhibit significant limitations. First, they focus on static graphs with fixed features, where network structures and node attributes remain unchanged throughout the learning process. Popular datasets like Cora, Citeseer, and Pubmed are commonly used for evaluation \cite{AL25, AL26, AL29}, but these lack the dynamic properties of real-world networks, where both graph structure and node features can evolve over time. These static evaluations fail to account for the evolving nature of real-world networks, limiting their applicability in adaptive settings like health monitoring, where data streams are continuous.

Furthermore, existing methods prioritize prediction accuracy but often neglect user/sensor burden, a critical aspect in real-world applications. Strategies that over-query central nodes or high-degree users can lead to user fatigue, sensor battery drain, and disengagement \cite{AL31}. Yet, user burden is rarely quantified or analyzed, limiting the sustainability of these methods in practical scenarios.  Third, existing AL approaches typically use single-time evaluations, failing to measure how model performance evolves over time. This ignores the critical aspect of model stability and adaptation in dynamic settings.

% Graph-based Active Learning (AL) offers a promising approach to address this challenge %by leveraging the structure of user networks to selectively query only a subset of users or sensors for data collection. 
% by using the graph structure and efficiently querying selection that maintains model performance while minimizing overall user burden.
% Despite these potential benefits, existing AL work on graph-based data has primarily focused on static datasets and denser networks, where network structures and data distributions remain fixed over time. %Common datasets used in such studies include Citeseer, Cora, and Pubmed for standard node classification benchmarks, along with denser networks like Corafull and Ogbn-Arxiv, and co-authorship networks. 
% While valuable for evaluating general model performance, these datasets lack the dynamic characteristics and real-life sensor data inherent in health monitoring and IoT applications. Furthermore, traditional graph AL approaches often prioritize prediction accuracy while overlooking practical concerns such as user burden, sampling diversity, and the risk of overburdening certain individuals within the network. These limitations restrict the applicability of existing AL methods in real world applications, where sustained user engagement are as essential as model performance.

To bridge this gap, we introduce GRAIL (GRaph ActIve Learning Benchmark), a systematic evaluation framework designed to assess graph-based AL strategies in time-series, stream-based settings that simulate real-world sensor networks. Unlike traditional AL methods that operate on static datasets, GRAIL integrates both spatial (graph) and temporal (time-series) dimensions, enabling a more accurate assessment of AL strategies under real-time, dynamic conditions. GRAIL is evaluated on two dynamic datasets (SNAPSHOT \cite{snapshot1} and FnF \cite{socialFMRI}), multi-modal sensor datasets reflecting complex, real-world social interactions and real-world variability.  Another novel aspect of GRAIL’s design is its rigorous approach to evaluation, which involves dividing nodes into distinct types (e.g., queried, unqueried, hold-out) to enable more granular and context-sensitive assessments of AL strategies across varied user/node profiles.

The benchmark introduces novel metrics, such as the Cumulative Performance Index (CPI), which evaluates the model’s sustained effectiveness over time, and sampling diversity metrics that track the distribution of queries across users, helping to identify patterns of over-exertion on high-centrality nodes. By examining each strategy’s robustness across various node types, from newly queried nodes to long-held-out test sets, the framework provides a holistic view of AL strategy performance in graph-based, health monitoring or IoT applications. %Furthermore, while this benchmark is designed with health applications in mind, it is built to be adaptable, making it a versatile tool that can be applied to evaluate AL strategies across diverse datasets and contexts.

The benchmarking framework evaluates 11 AL strategies along three key dimensions:  
\begin{itemize}
   
 \item Performance Metrics: Including accuracy, Area Under the Curve (AUC), and CPI to quantify the predictive capabilities of each strategy over time.  
 \item  Diversity and Burden Metrics: Measuring the degree of query diversity and potential user fatigue alongside time-gap analyses that reveal how frequently individual users are queried.  
 \item Network-Level Analysis: Examining the influence of network topology on strategy effectiveness, particularly focusing on centrality and exertion patterns across users.

\end{itemize}
Findings from these experiments reveal critical insights into the trade-offs AL strategies must manage between prediction performance and user burden. They also highlight key limitations in existing approaches, inspiring future development of AL methods that are sustainable, efficient, and mindful of the user experience in dynamic health and well-being monitoring systems. By providing a structured, reproducible evaluation framework, this benchmark serves as a foundational resource for advancing AL strategies across multiple domains and datasets, supporting the development of adaptable, real-time applications. The code and dataset are available at Github and Kaggle \cite{GRAIL_Code_2025, GRAIL_Dataset_2025}.

\section{GRAIL: Graph-Based Active Learning Framework for Dynamic Sensing Data}

\begin{figure}
    \centering
    \includegraphics[width=0.99\linewidth]{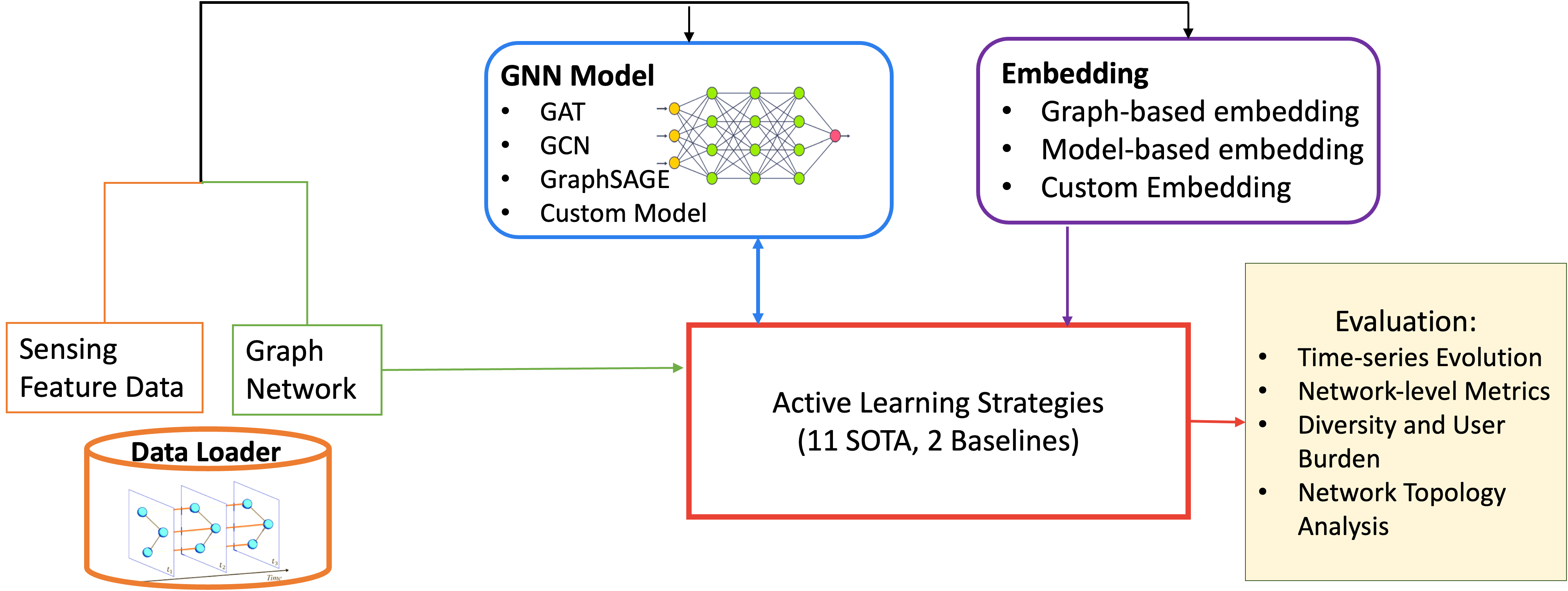}
    \caption{GRAIL Framework for Graph Active Learning in Dynamic Sensing Environments}
    \label{fig:GRAIL_framework}
\end{figure}

GRAIL’s core is composed of five flexible components: Data Loader, GNN Models, Embedding module,  AL Strategies, and multi-dimensional Evaluation module.
The framework is shown in Figure. \ref{fig:GRAIL_framework}.

\begin{itemize}
\item \textbf{Data Loader}
The data loader provides options to load processed data, including features and graphs, for multiple datasets with various labels and graph types (e.g., call, SMS, friend networks). Examples include SNAPSHOT and FnF (with labels such as mood or sleep).

    \item \textbf{Model Options}: GRAIL supports various Graph Neural Network (GNN) architectures, including GAT (Graph Attention Network)\cite{gat_layer}, SAGE GNN (GraphSAGE)\cite{sage_layer}, GCN (Graph Convolutional Network)\cite{gcn_layer}, and allows for Custom Models. This adaptability accommodates different data and task requirements, allowing users to tailor the framework to their needs.

    \item \textbf{Active Learning (AL) Strategy Options}: GRAIL includes 11 AL strategies for selectively querying nodes which can be broadly categorized into four main groups: %(1) Baseline methods, including No Active Learning and Random Sampling; 
    (1) Uncertainty-based sampling techniques\cite{AL_entropy, AL25}; (2) Graph-structure-based sampling methods, including Degree-Based and PageRank-Based sampling \cite{AL25, AL26}; (3) Embedding-space clustering methods such as dentisy-based, coreset, FeatProp\cite{AL26, AL_coreset, AL27}  and (4) Hybrid strategies %that combine graph structure, input features, and model outputs, 
    such as GraphPart, GraphPartFar, and Active Graph Embedding (AGE) \cite{AL28, AL26}.
    % Available strategies encompass Uncertainty-based \textcolor{red}{Maryam, please check here} (\cite{AL_entropy,AL25}), Diversity-based, and Graph-based approaches (degree-based \cite{AL25}, pagerank-based \cite{AL26} alongside Hybrid Methods (GraphPar, GraphPartFar\cite{AL28}, AGE \cite{AL26}) that combine multiple strategies and 
    Details are provided in Appendix Table \ref{tab:strategy_summary}. Users can also implement Custom Strategies to evaluate novel methods suited to specific contexts. 

    \item \textbf{Embedding Options}: For embedding computation, GRAIL offers two distinct approaches:
    \begin{itemize}
        \item Model-based: Utilizing the trained GNN model to generate embeddings, leveraging learned representations from the trained model for more integrated, task-specific embeddings.
        \item Direct GNN: Applying a separate GNN model to pass data and graph structure through its layers, obtaining embeddings independent of the trained model, thus allowing flexibility in embedding generation separate from specific model tasks or training phases.
    \end{itemize}

\item \textbf{Evaluation Module}: The GRAIL framework features a comprehensive evaluation pipeline, assessing AL strategies along three critical dimensions: Model Performance, Sampling Analysis, and Network Analysis.

\begin{itemize}
    \item \textbf{Model Performance (Section 3)}: This component includes metrics such as Accuracy, F1 Score, AUC, and Cumulative Performance Index (CPI), which assess the predictive strength of the model across dynamic, real-world conditions. In addition to the custom metrics like CPI, users can select from up to seven metrics: accuracy, micro F1, macro F1, recall, precision, Area Under the ROC Curve (AUC-ROC), and Area Under the Precision-Recall Curve (AUC-PR).

    \item \textbf{Sampling Analysis (Section 4)}: GRAIL evaluates the diversity and distribution of queries with metrics like Diversity Metrics (measuring range and query balance), Coverage Ratio (extent of node sampling), Time Distribution (frequency of queries over time), and Burden Metrics (assessing user query fatigue).
    
    \item \textbf{Network Analysis (Section 4)}: To assess how AL strategies perform across node types, GRAIL includes Node Types analysis (e.g., high-centrality vs. peripheral nodes), Time-Gap Analysis (intervals between repeated queries on nodes), Centrality Impact (influence of high-centrality nodes on sampling), and Graph Properties (such as density and clustering, affecting AL effectiveness).
\end{itemize}

\end{itemize}

\section{Network-level Performance Evaluation for Dynamic Graph Systems}

This section outlines the process of generating metrics for a Graph Neural Networks (GNN) model using time-series data. The experiment focuses on the model's performance across different types of nodes (train, unqueried, same-day unqueried, holdout test) and employs bootstrapping to ensure robust evaluation. Key design parameters such as the number of initial training days (L), the number of selected nodes (k), and the ratio of selected nodes to total nodes (k/N) are incorporated into the evaluation.

\subsection{Categorizing Nodes for Evaluation}

\begin{figure}
    \centering
    \includegraphics[width=0.8\linewidth]{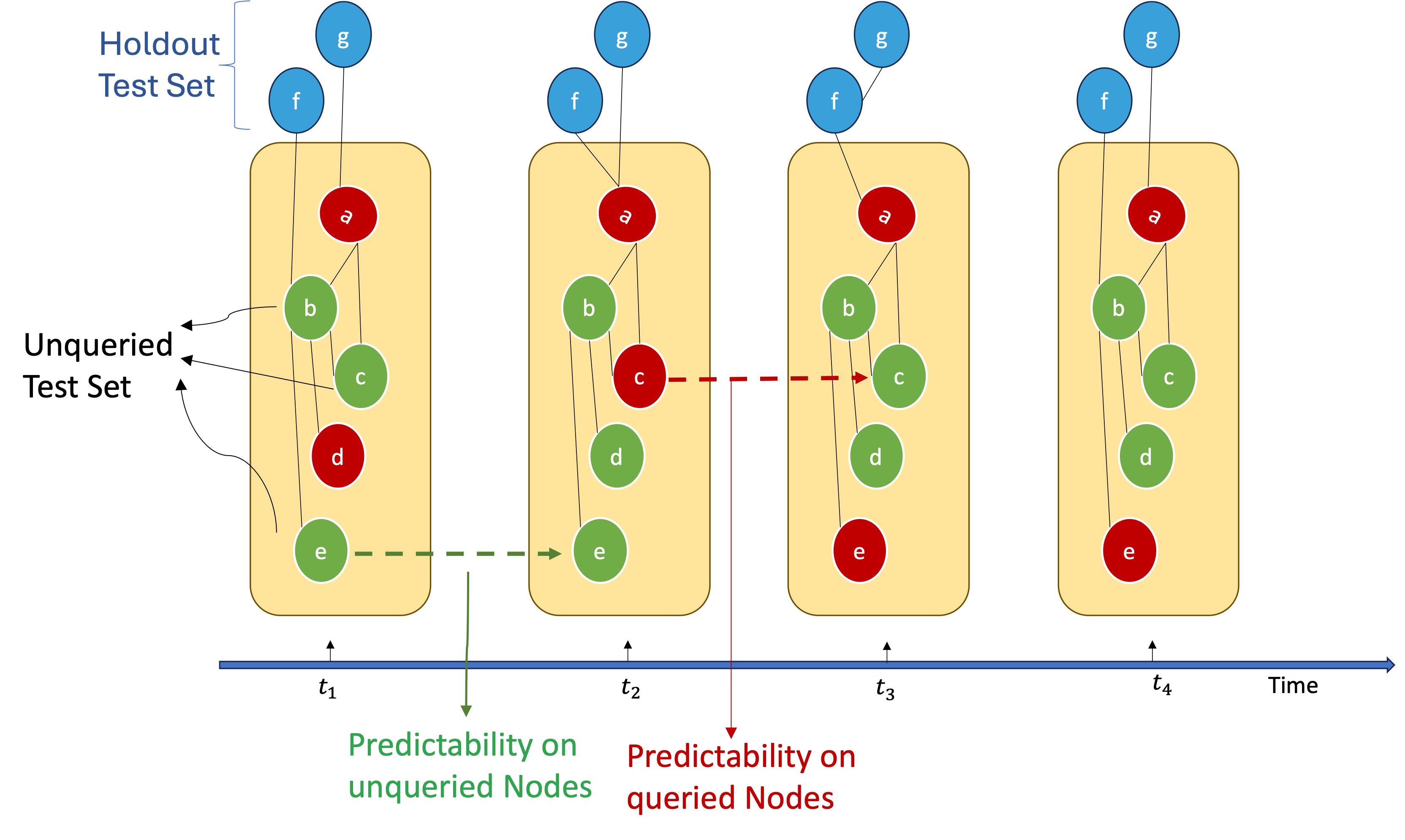}
    \caption{Evaluation Strategy overview of how nodes are categorized and divided over time. Timepoints represent instances where the network state is assessed and nodes are queried. Blue nodes are part of the holdout set that remains unseen by the model. The yellow box indicates the query pool. Red nodes represent queried nodes, while green nodes are unqueried. }
    \label{fig:enter-label}
\end{figure}
Given the model's reliance on spatial (graph) and temporal (time-series) components, an effective evaluation strategy is necessary to differentiate the model’s predictive performance across various node categories. This ensures a comprehensive understanding of how the model generalizes beyond its training set, particularly when predicting outcomes for unqueried or unseen nodes. The experiment is designed to answer questions such as: How well does the model perform on unqueried nodes when k nodes have been queried from the query pool? The node categories for each query timepoint (e.g., each day) in the time series are explained below and in Figure \ref{fig:enter-label}:

\begin{itemize}
    \item \textbf{Train Nodes}: These nodes are used for training the model on each day of the time series. Metrics are computed based on the model's predictions for these nodes, providing insight into how well the model is fitting the training data.
    
    \item \textbf{Unqueried Nodes}: These nodes are excluded from the training process on each specific day. The model's predictions on these unqueried nodes help assess how well it generalizes to unseen data.
    
    \item \textbf{Same-Day Unqueried Nodes}: These are the nodes that are not used for training on the same day but are included for predictions on subsequent days. This category provides insight into the model’s performance on unqueried nodes but is later predicted in the next step.

    \item \textbf{Holdout Test Set}: This set consists of nodes that are never used for training or querying throughout the experiment. Metrics from the holdout set provide a benchmark for the model's ability to predict on an entirely unseen data, showcasing its overall generalization capability and robustness. 
\end{itemize}
Dividing nodes into these four categories is crucial for stream AL as it provides a comprehensive view of the model's performance under varied data exposure. Distinguishing between queried and unqueried nodes helps evaluate the effectiveness of AL strategies where only a subset of nodes is queried. This analysis reveals how well the model can leverage limited labeled data to make accurate predictions on unseen nodes.

Assessing same-day unqueried nodes highlights the model’s immediate predictive capabilities and adaptability to changing data. Evaluating the holdout test set ensures that the model's performance generalizes to entirely unseen data, essential for real-world scenarios where future data predictions are required. This differentiation aids in refining AL strategies, prioritizing nodes that maximize predictive improvements, and optimizing both model accuracy and resource use.

\subsection{Performance Metrics}
The following metrics are generated during each bootstrap iteration and across the three node categories:We evaluated model performance in each bootstrap iteration and across the three node categories using the following metrics: Accuracy, AUC ROC, AUC Precision-Recall, Precision, and Recall. These standard metrics provide a comprehensive view of classification performance.

Our key contribution is the Cumulative Performance Index (CPI), a novel metric designed to capture sustained model effectiveness over time in a dynamic AL setting. Unlike standard metrics that assess static performance, CPI quantifies a model's performance trajectory across multiple timepoints.
   We utilize AUC, which is a metric used to represent the cumulative performance of a model over multiple time points in AL experiments. As the AL cycle progresses, queries are made at each timepoint, and model performance (e.g., F1 score, accuracy, or any other relevant metric that takes values between 0 and 1) is evaluated on the unqueried nodes or future timepoints. The AUC is computed by summing the performance values across all timepoints, capturing the overall model effectiveness over time.

Formally, the AUC can be represented as:
\[
\text{AUC} = \sum_{t=1}^{T} \frac{\text{Performance}(t) + \text{Performance}(t+1)}{2} \times \Delta t
\]
where \( T \) is the total number of timepoints, and \( \Delta t \) is the interval between timepoints (e.g., 1 hour or 1 day). This trapezoidal approximation integrates the performance curve over the specified timepoints.
We define the CPI by normalizing the AUC. The CPI is obtained by dividing the AUC by \( T \), yielding a value between 0 and 1:
\[
\text{CPI} = \frac{\text{AUC}}{T}
\]

The maximum value of CPI is 1, which is when the model provides 100\% performance at all time points.
This CPI metric provides a standardized way to compare the effectiveness of AL strategies across various experimental setups, regardless of the number of timepoints.

\section{Diversity \& Sampling Burden Metrics for Dynamic Graph Systems}
In dynamic graph-based environments (e.g., social networks, sensor networks, IoT), AL can impose significant query burden on specific nodes. Even with static graph structures, node features (e.g., sensor data) can change dynamically.  We introduce and evaluate a set of metrics designed to assess AL strategies on two key aspects: sampling diversity and user burden. These metrics allow us to quantify how well an AL strategy distributes queries across nodes, avoids over-exertion, and maintains a fair representation of the network. 
\begin{itemize}
\item \textbf{Sampling Entropy:}It measures the diversity in node selection by quantifying how evenly nodes are sampled over the course of the AL process. Higher entropy values indicate a more uniform distribution, meaning that the AL strategy is querying a broad set of nodes rather than focusing on a few. Entropy is computed as:
\begin{equation}
    H = - \sum_{i=1}^{N} p_i \log(p_i)
\end{equation}
where $p_i$ is the probability of sampling node $i$, and $N$ is the total number of nodes.

\item \textbf{Coverage Ratio:} It assesses the proportion of nodes that are sampled at least once over the evaluation period. It ensures that the AL strategy is not only sampling a few key nodes but is capturing a broader section of the graph. The coverage ratio is given by:
\begin{equation}
    C = \frac{\text{Number of unique nodes sampled}}{\text{Total number of nodes (N)}}
\end{equation}

\item \textbf{Time-Gap Analysis} evaluates how frequently the same nodes are being sampled across consecutive or nearby days. This metric helps identify over-exertion, where certain nodes are queried too frequently, leading to fatigue (for users) or battery depletion (for sensors).

\begin{itemize}
    \item \textbf{Average Time Gap}: This metric calculates the mean number of days between consecutive samples for each node:
    \begin{equation}
        \text{Average Time Gap} = \frac{1}{N} \sum_{i=1}^{N} \frac{1}{|T_i|} \sum_{t=1}^{|T_i| - 1} (T_{i,t+1} - T_{i,t})
    \end{equation}
    where $T_i$ is the set of days when node $i$ is sampled, and $N$ is the total number of nodes.
    
    \item \textbf{Within-Gap Percentage}: This metric computes the percentage of nodes sampled within a specified gap (e.g., 1 or 2 days):
    \begin{equation}
        \text{Within-Gap Percentage} = \frac{\sum_{i=1}^{N} \mathbb{1}(\text{gap} < k)}{N}
    \end{equation}
    where $k$ is the specified time gap threshold and $\mathbb{1}$ is an indicator function.
    
    \item \textbf{Over-Exertion Score}: The \textbf{over-exertion score} quantifies how often nodes are sampled more frequently than a defined time-gap threshold. High over-exertion scores indicate that certain nodes are being repeatedly sampled within short intervals:
    \begin{equation}
        E = \frac{\text{Number of nodes sampled within the gap threshold}}{\text{Total number of nodes sampled}}
    \end{equation}
\end{itemize}

\end{itemize}
\subsection{Centrality Metrics}
In graph-based AL strategies, nodes have varying levels of importance based on their position in the graph. Centrality metrics help us understand which nodes are more critical in the network and how their centrality correlates with their sampling frequency. Key centrality metrics included in GRAIL are Degree Centrality, Betweenness Centrality, Closeness Centrality, Eigenvector Centrality, Harmonic, Load, PageRank, and Clustering coefficient (details in appendix \ref{centrality_def}). By analyzing centrality metrics alongside sampling and time-gap statistics, we can determine whether certain high-centrality nodes are being disproportionately favored by the AL strategy. 

% \subsection{Bootstrapping Across Days}
% Bootstrapping is used to simulate multiple independent trials of the AL process over several days. This method allows us to assess the variability in node selection across different runs, making the results more robust to fluctuations in the graph or user behavior. Each bootstrap trial represents a fresh sampling run, with diversity and exertion metrics computed for each day to capture daily variability. 

% By analyzing these metrics over various bootstrap runs, we can evaluate how consistent the AL strategy is in spreading the querying burden, ensuring it does not repeatedly stress certain nodes while maintaining broad and informative data collection.

\section{Datasets}
We used two dynamic social network datasets:  SNAPSHOT\cite{snapshot1} and Friends-and-Family (FnF)\cite{socialFMRI}. SNAPSHOT, in processed form, is a publicly available dataset from a longitudinal study with 200+ participants using wearable and mobile sensors (e.g., skin conductance, acceleration, skin temperature, SMS and calls meta data, screen use, mobility patterns), some of whom are also socially connected. FnF, a similar study with 130 participants, collected raw sensor data (e.g., calls, SMS, Bluetooth, location) over 15 months, but is only available in raw form. We developed a preprocessing pipeline for FnF, including data cleaning, feature engineering, and temporal alignment, transforming it into a structured graph format suitable for Active Learning (AL).

For both datasets, graph networks were constructed using SMS/Call meta data. Additionally, the FnF dataset included a friendship graph derived from survey data. All these graphs are provided in the released dataset for experimentation. The results reported here specifically correspond to the SMS Graph for SNAPSHOT and the Friendship Graph for FnF.  Model inputs consist of sensor data, while model outputs are binary labels for mood and sleep duration. The dataset is divided into a held-out set (20\% of the data) and an active learning (AL) pool (80\% of the data). During each AL iteration, $k$ nodes are sampled from the pool. The model is initially trained using data from the first $L$ days. See dataset and processing details in the appendix \ref{Dataset_details}.

% Graph-based AL on dynamic, real-world sensor data is rare. Existing benchmarks use static graphs or simulated data, lacking the temporal complexity and sensor-driven interactions of real-world scenarios. Our work bridges this gap, providing a benchmark for AL on two dynamic graph datasets. Details of SNAPSHOT and FnF preprocessing are provided in the appendix \ref{Dataset_details}.
% % \subsection{Dataset: SNAPSHOT}

% \subsection{Dataset: Friends and Family}

% The Friends and Family dataset offers a comprehensive, multi-modal collection of data focusing on social behavior and decision-making within a residential community adjacent to a major research university in North America\cite{socialFMRI}. The study was conducted in two phases. \textbf{Phase I}, starting in Spring 2010, included 55 participants, while \textbf{Phase II}, launched in Fall 2010, expanded the study to include 130 participants across approximately 64 families. Data collection was conducted over a 15-month period using Android-based mobile phones equipped with various sensors to track the subjects’ activities and interactions.

% The dataset contains detailed logs of multiple types of sensor data and survey responses, each providing insights into participants’ activities, interactions, and social networks (See details in Supplementary Materials). 

%\section{Proposed Approach}
%Our approach leverages graph active learning to optimize data collection strategies by minimizing the user burden while ensuring data relevance...

\section{Experiments and Results}

% \textcolor{red}{Maryam, can you add one sentence about which machine, which GPU you used for running this experiment?}\textcolor{blue}{yes I'll add it. Btw I have shortened the paper to 9 pages. Now I will add some results to supplementary and then move to code}

% \textcolor{red}{Got it. can you also go through the checklist? I will add some more comments}\textcolor{blue}{sure, let me do it after code. }
% % \textcolor{red}{The experimental setting should be presented in the core of the paper to a level of detail
% that is necessary to appreciate the results and make sense of them. The full details can be provided either with the code, in appendix, or as supplemental
% material}

% \textcolor{red}{ provide sufficient information on the computer resources (type of compute workers, memory, time of execution) needed to reproduce the experiments}
GRAIL is designed with a flexible and systematic structure, allowing users to tailor experimental settings based on their application needs. Key experiment design features include:
\begin{itemize}
    \item Training Period and Query Size (\( L, k \)): \( L \) represents the number of initial timepoints (days in our evaluation) for which all nodes are used for training, providing a stable initial model. \( k \) defines the number of nodes queried for labels each day after the initial training period, controlling the level of query burden. 
    
    \item Bootstrapping:To ensure a robust performance evaluation, a bootstrapping procedure(set to 100 iterations) is employed. Each bootstrap iteration spans one AL cycle for a given AL strategy and chosen set of parameters with random initialization. 
    \item Time Series/Stream Evaluation: The experiment is conducted at a day-level where, each day, a subset of nodes is queried, the model is updated, and performance is evaluated. This setup captures temporal variations in model performance, making it well-suited for real-world, stream-based applications.

\end{itemize}

We conducted two major sets of experiments to comprehensively evaluate the effectiveness of AL strategies for dynamic social network data.

\textbf{(1) Fine-grained, dynamic analysis of AL strategies: }
The first set of experiments focuses on fine-grained, dynamic analysis of AL strategies within each dataset individually. Specifically, it explores how each strategy performs over time, offering a detailed view of temporal variation in performance and sampling behavior. We analyzed how each AL strategy's performance varies daily, using metrics like CPI, Accuracy, and F1 Macro. Temporal trends were captured using rolling averages, revealing patterns such as learning stability, gradual improvement, or sudden drops in performance. We also assessed how AL strategies balance diversity and user burden across days, using metrics like Sampling Entropy, Coverage Ratio, and Over-Exertion Scores. The analysis highlights how AL strategies adapt to the dynamic user behavior and network structure in the SNAPSHOT and FnF datasets, making the insights dataset-specific. Our findings show that strategies like GraphPart and GraphPartFar consistently demonstrated high performance across various node categories, while strategies such as coreset and uncertainty-based methods maintained balanced sampling without overburdening central nodes. Due to the extensive length of this analysis, complete details for these experiments are provided in the Supplementary Materials.

\textbf{(2) Performance-burden tradeoff analysis: }
The second set of experiments shifts focus to a comparative, time-independent analysis, where we directly compared the performance-burden tradeoff of all AL strategies across both datasets. Instead of tracking daily performance, this analysis aggregates results across all time points, providing a stable view of each strategy's average performance and burden. Our analysis focuses on two core aspects: predictive performance vs user burden tradeoff and impact of network topology on user burden.

\begin{figure}[htbp]
    \centering
    \begin{minipage}[t]{0.475\textwidth}
        \centering
        \includegraphics[width=\linewidth]{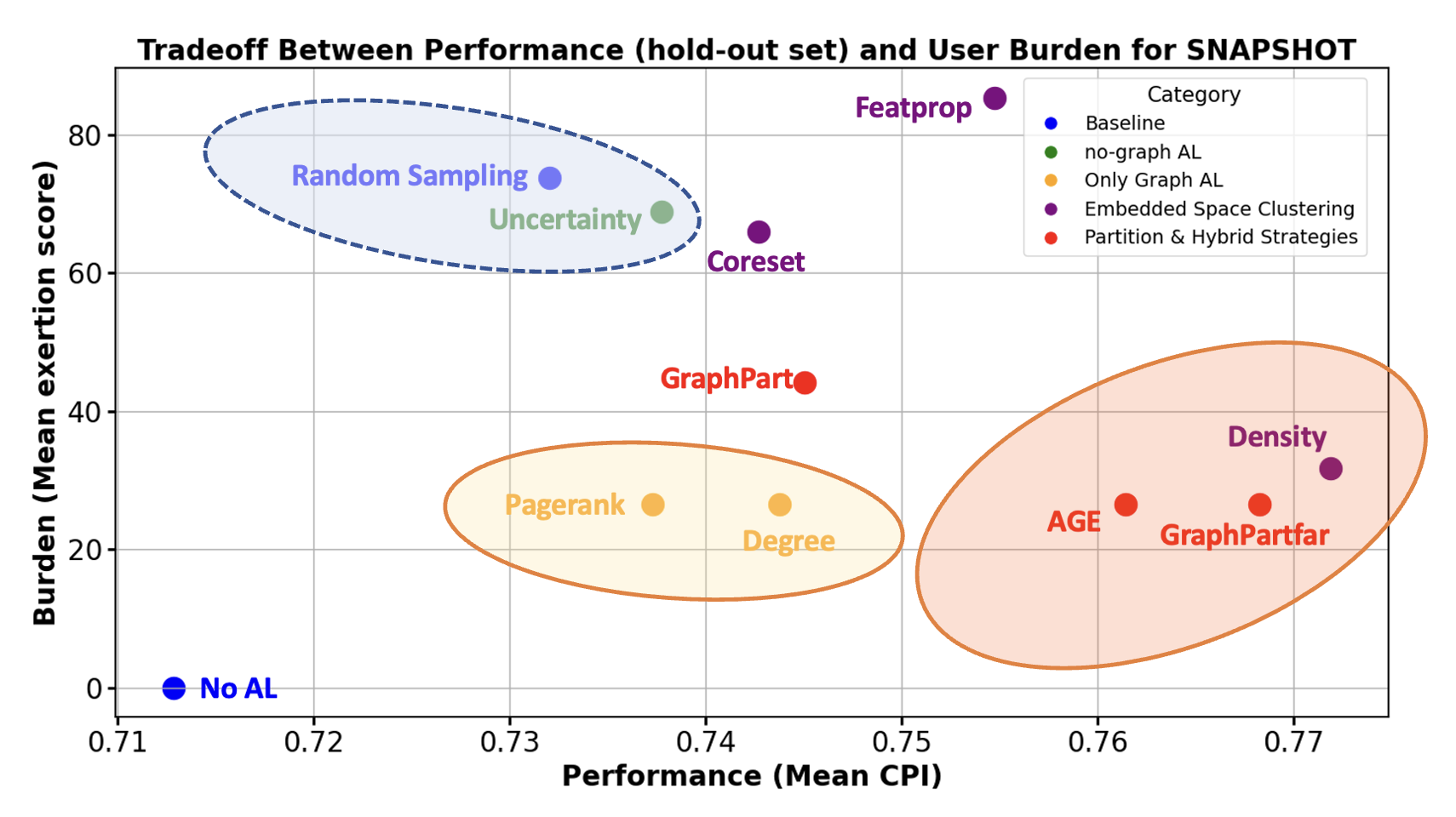}
        \caption{Mean CPI vs Mean Exertion Score for SNAPSHOT sleep classification for Cohort 2 using the SMS Graph (L=8, k=9, gap=5).}
        \label{fig:snap_tt}
    \end{minipage}\hfill
    \begin{minipage}[t]{0.475\textwidth}
        \centering
        \includegraphics[width=\linewidth]{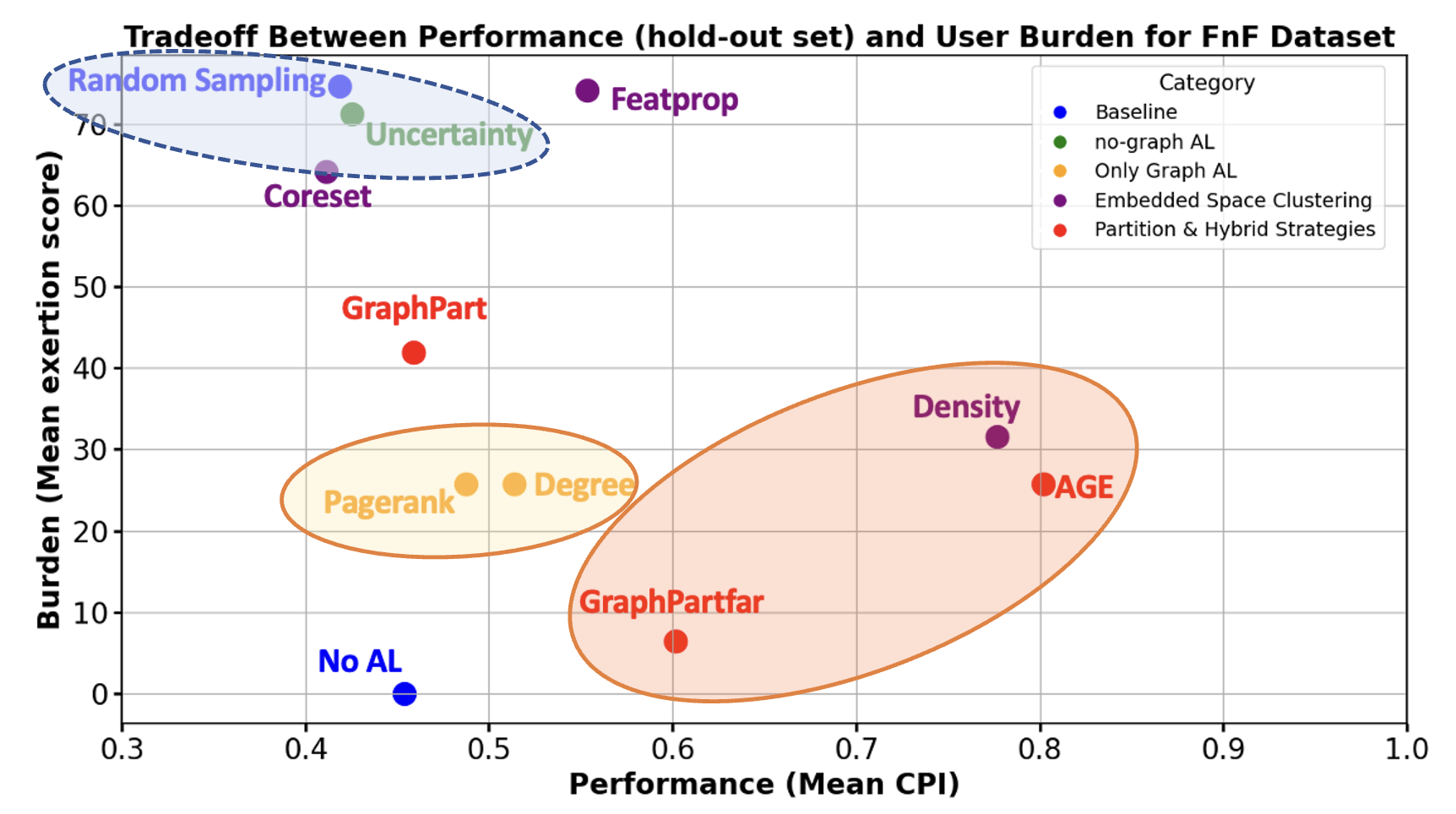}
        \caption{Mean CPI vs Mean Exertion Score for FnF sleep classification using the Friendship Graph (L=6, k=8, gap=5).}
        \label{fig:fnf_tt}
    \end{minipage}
\end{figure}

\begin{figure}[htbp]
    \centering
   
    \begin{minipage}[t]{0.47\textwidth}
        \centering
        \includegraphics[width=\linewidth]{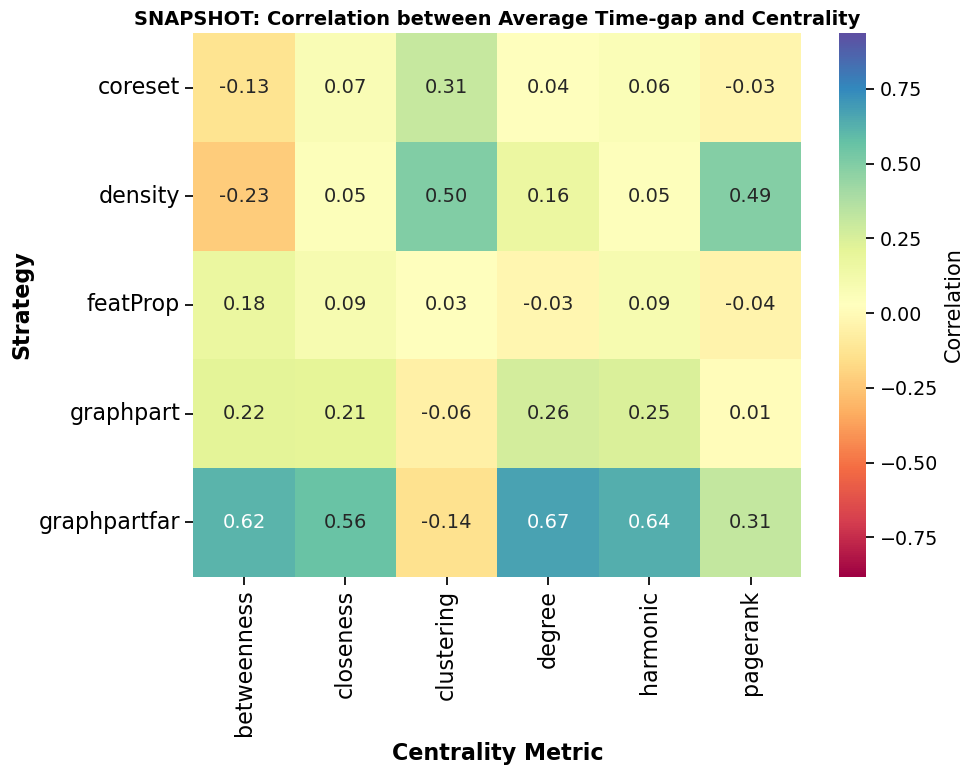}
        \caption{Correlation between centralities and burden for SNAPSHOT sleep classification for Cohort 2 using the SMS Graph (L=8, k=9, gap=5).}
        \label{fig:snap_corr}
    \end{minipage}\hfill
     \begin{minipage}[t]{0.47\textwidth}
        \centering
        \includegraphics[width=\linewidth]{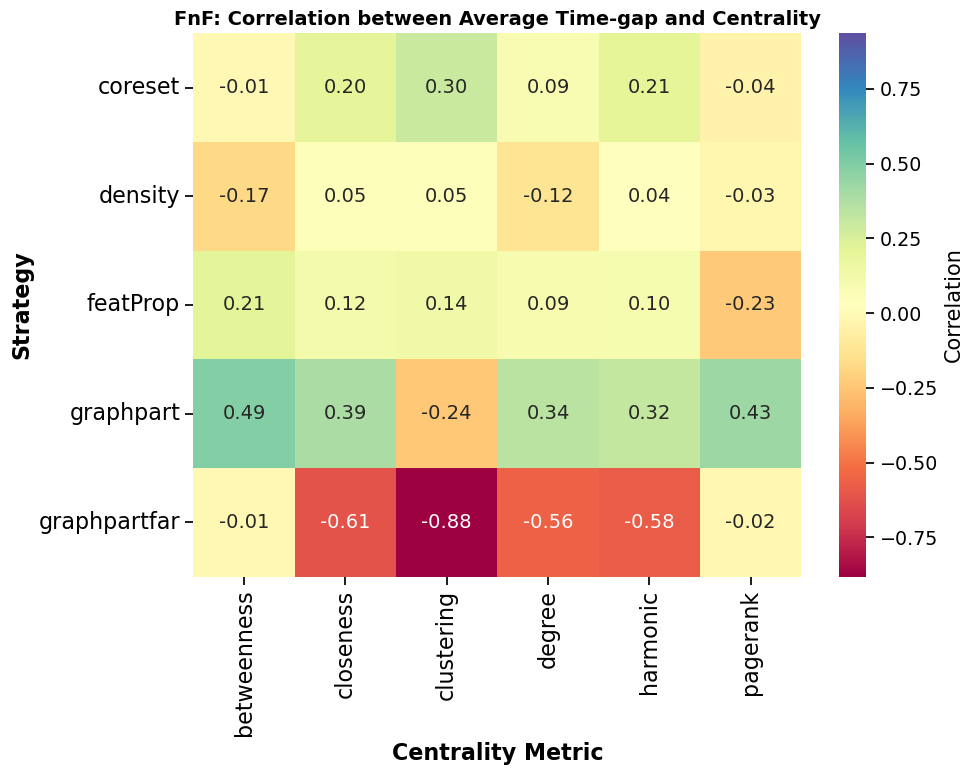}
        \caption{Correlation between centralities and burden for FnF sleep classification using the Friendship Graph (L=6, k=8, gap=5).}
        \label{fig:fnf_corr}
    \end{minipage}
    
\end{figure}
% \begin{figure}[htbp]
%     \centering
%     \begin{subfigure}[t]{\textwidth}
%         \centering
%         \includegraphics[width=0.8\textwidth]{Figures/SS/fnf_corr.png} % Replace with your second image file
%         \caption{Correlation between centralities and burden for FnF sleep classification using the Friendship Graph (L=6, k=8, gap=5).}
%         \label{fig:fnf_corr}
%     \end{subfigure}
%      \vspace{0.5cm} 
%     \begin{subfigure}[t]{\textwidth}
%         \centering
%         \includegraphics[width=0.8\textwidth]{Figures/SS/snap_corr.png} % 
%         \caption{ Correlation between centralities and burden for SNAPSHOT sleep classification for Cohort 2 using the SMS Graph (L=8, k=9, gap=5).}
%         \label{fig:snap_corr}
%     \end{subfigure}
%     \vspace{0.5cm} 
%     \caption{Correlation between centralities and burden for sleep classification on the hold-out set with different datasets.}
%     \label{fig:main_corr}
% \end{figure}

\begin{itemize}
    \item \textbf{Predictive Performance:}
Figures \ref{fig:snap_tt} and \ref{fig:fnf_tt} reveal critical differences in the predictive capabilities of the AL strategies across two datasets. The ideal
strategy should have the highest CPI and the lowest burden. In the SNAPSHOT dataset, No AL maintains the lowest burden but offers moderate performance, while Random Sampling increases burden without notable gains. Uncertainty-based AL (no-graph) slightly improves performance over Random Sampling but remains in a high-burden cluster. Graph-based strategies like AGE and GraphPartfar achieve high performance with moderate burden. Density and GraphPart further push performance at a higher user burden, while Pagerank and Degree show moderate performance.

In contrast, the FnF dataset shows a similar pattern but with higher variability in user burden among graph-based strategies. AGE and GraphPartfar form a high-performance cluster, but the dispersion is more pronounced due to the decentralized topology. Density and GraphPart deliver strong performance at a higher user burden, while Pagerank and Degree maintain moderate performance.

Across both datasets, AGE demonstrates the best trade-off between performance and burden, balancing centrality, uncertainty, and density. The observed differences in cluster dispersion between the two datasets highlight the impact of network topology, which we further explain by analyzing network topology and centrality correlation.

  \item \textbf{Impact of Network Topology and User Burden:} The correlation between burden and centralities in Figure \ref{fig:snap_corr} and  \ref{fig:fnf_corr} and network centralities distribution in Figure \ref{fig:topology} (in appendix) collectively provide valuable insights into the interplay between graph topologies, centrality distributions, and the behavior of sampling strategies.
  
In SNAPSHOT, strategies like GraphPartfar show positive correlations with centrality metrics (betweenness, closeness, degree), maintaining sufficient time gaps between central nodes and achieving balanced sampling. Conversely, in FnF, GraphPartfar and GraphPart show negative correlations, frequently sampling central nodes due to the network's decentralized structure with diverse community sizes. This results in over-exertion and reduced time gaps.

The tradeoff Figures \ref{fig:snap_tt} and \ref{fig:fnf_tt} further support these observations. In SNAPSHOT, strategies like AGE and GraphPartfar form a high-performance, low-burden cluster, leveraging a uniform topology. In FnF, the same strategies exhibit higher variability, reflecting challenges from its heterogeneous structure. FeatProp and Coreset, less influenced by topology, maintain consistent performance across both datasets.

Overall, these findings emphasize the importance of aligning AL strategies with graph topology. Uniform networks like SNAPSHOT favor balanced sampling, while decentralized structures like FnF require careful tuning to avoid over-exertion of central nodes.

\end{itemize}
In summary, our results demonstrate that the choice of AL strategy must account for network topology and user burden. AGE offers the most consistent performance–burden trade-off across datasets, while GraphPartFar excels in predictive performance at the cost of higher user burden. These insights validate the importance of considering both predictive performance and user engagement in dynamic graph AL settings.

The experiments were run on an AMD Ryzen Threadripper 3970X (32 cores, 3.7 GHz), two RTX 3090 GPUs (24 GB of VRAM each) and 251 GiB RAM on Ubuntu 20.04.1. An experiment with one dataset, fixed parameters, and 100 bootstraps took around 30 minutes.

\section{Limitations}

Although this study presents significant advances in AL for dynamic social networks, there are several limitations primarily arising from the current state of available datasets and technological constraints.

\begin{itemize}
    \item \textbf{Dataset Diversity and Static Graphs:} Our analysis relies on two dynamic social network datasets (SNAPSHOT and FnF), primarily focused on health monitoring with sensor data. Although GRAIL is designed for dynamic graphs, we used static graph structures with evolving sensor data due to the lack of datasets featuring dynamic graph structures. The available datasets also capture relatively simple social structures, lacking the complexity of larger networks.

    \item \textbf{Sample Size and Recruitment Constraints:} The datasets used have limited participant numbers (200+ for SNAPSHOT and 130 for FnF) due to the high user burden in health sensing studies. Recruiting socially connected participants further restricted dataset size.

    \item \textbf{Scalability and Larger Networks:} GRAIL's scalability was not fully tested on larger, complex networks due to the absence of suitable datasets, limiting our ability to assess AL strategy performance in such scenarios.

    \item \textbf{Fixed Querying Intervals:} Our experiments employed a uniform daily query interval, which may not be optimal for user engagement or model performance. Adaptive query intervals based on user behavior or model uncertainty could be explored in future work.

    \item \textbf{Simplified User Burden Measurement:} User burden was measured primarily by sampling frequency and time-gap, overlooking cognitive load, availability, or context. A multi-dimensional burden model could provide a more user-centric evaluation.

    \item \textbf{Custom Preprocessing for Specific Datasets:} The preprocessing pipeline was tailored to SNAPSHOT and FnF, ensuring data quality but limiting direct applicability to other datasets without adaptation. However, GRAIL’s modular design allows adjustments for other domains.
\end{itemize}

\section{Conclusion}
We introduced GRAIL, a benchmark for graph AL on dynamic, time-series data, revealing how graph-based strategies like GraphPartFar and AGE achieve superior performance while balancing user burden. Our results highlight the critical impact of network topology on strategy effectiveness, and our proposed metrics (CPI, entropy, coverage) provide a robust foundation for evaluating graph AL methods. GRAIL is publicly available, allowing researchers to compare and identify the most suitable AL strategies for their own dynamic graph applications.

\bibliography{references}
\bibliographystyle{icml2021}

%%%%%%%%%%%%%%%%%%%%%%%%%%%%%%%%%%%%%%%%%%%%%%%%%%%%%%%%%%%%%%%%%%%%%%%%%%%%%%%
%%%%%%%%%%%%%%%%%%%%%%%%%%%%%%%%%%%%%%%%%%%%%%%%%%%%%%%%%%%%%%%%%%%%%%%%%%%%%%%
% DELETE THIS PART. DO NOT PLACE CONTENT AFTER THE REFERENCES!
%%%%%%%%%%%%%%%%%%%%%%%%%%%%%%%%%%%%%%%%%%%%%%%%%%%%%%%%%%%%%%%%%%%%%%%%%%%%%%%
%%%%%%%%%%%%%%%%%%%%%%%%%%%%%%%%%%%%%%%%%%%%%%%%%%%%%%%%%%%%%%%%%%%%%%%%%%%%%%%

\newpage
\appendix
\section{Technical Appendices and Supplementary Material}

\subsection{Active Learning (AL) Strategies}
The existing state-of-the-art AL strategies are listed in Table \ref{tab:strategy_summary}.
\begin{table*}

\centering
% \setstretch{1.2} % Adjust line spacing within cells
\resizebox{0.9\textwidth}{!}{%
\begin{tabular}{|p{3.8cm}|p{12cm}|}
\hline
\multicolumn{2}{|c|}{\textbf{Summary of Active Learning Strategies}} \\ \hline
\multicolumn{2}{|l|}{\textbf{Baseline}} \\ \hline
\textbf{Strategy} & \textbf{Description} \\ \hline
No Active Learning (AL) & The model is trained on the first L days of data but not updated with new queried labels. Serves as a baseline without any active learning interventions. \\ \hline
Random Sampling & Randomly selects nodes without considering features, predictions, or graph structure. Serves as a simple baseline. \\ \hline
\multicolumn{2}{|l|}{\textbf{Uncertainty-Based Sampling (No Graph Use)}} \\ \hline
Entropy-Based Uncertainty\cite{AL_entropy} & Selects nodes with the highest entropy in their predicted probabilities. Focuses on instances with evenly distributed probabilities indicating maximum uncertainty. \\ \hline
Least Confidence Uncertainty\cite{AL25} & Targets nodes with the lowest maximum predicted class probability. Identifies cases where the model is least confident about its predictions. \\ \hline
Max Margin Uncertainty\cite{AL25} & Selects nodes with the smallest difference (margin) between the top two predicted class probabilities. Focuses on the most ambiguous cases. \\ \hline
\multicolumn{2}{|l|}{\textbf{Only Graph-Structure-Based Sampling}} \\ \hline
Degree-Based Sampling\cite{AL25} & Uses graph structure to select nodes with the highest degree centrality. These nodes are highly connected and influential within the network. \\ \hline
PageRank-Based Sampling\cite{AL26} & Selects nodes with the highest PageRank centrality, prioritizing those that are influential in the graph structure. \\ \hline
\multicolumn{2}{|l|}{\textbf{Embedding-Space Clustering (Graph+Input Features)}} \\ \hline
Density-Based Sampling\cite{AL26} & Clusters nodes in the embedding space and selects those closest to cluster centers. Ensures representation of dense regions. \\ \hline
Coreset Sampling\cite{AL_coreset} & Employs K-Center clustering to maximize coverage in the embedding space. Selected nodes are spread out to reduce redundancy and ensure diversity. \\ \hline
FeatProp (Feature Propagation)\cite{AL27} & Propagates features through the graph to minimize classification error bounds. Selects nodes closest to cluster centers in the embedding space. \\ \hline
\multicolumn{2}{|l|}{\textbf{Partitioning and Hybrid Strategies (Graph + Input Features + Model)}} \\ \hline
GraphPart\cite{AL28} & Partitions the graph into disjoint communities using modularity-based clustering and selects representative nodes from each partition using K-Medoids. Balances representation across regions. \\ \hline
GraphPartFar\cite{AL28} & Extends GraphPart by penalizing nodes close to previously selected ones. Ensures diversity and broader graph coverage. \\ \hline
AGE (Active Graph Embedding)\cite{AL26} & Combines clustering, network structure (e.g., centrality), and uncertainty metrics (e.g., entropy). \\ \hline
\end{tabular}%
}
\caption{Summary of Active Learning Strategies}
\label{tab:strategy_summary}
\end{table*}

\subsection{Centrality Definitions}
\label{centrality_def}
\begin{itemize}
    \item \textbf{Degree Centrality}: Measures the number of direct connections that a node has. A high degree centrality indicates a well-connected node.
    \[
    \text{Degree Centrality}(v) = \frac{\text{deg}(v)}{N-1}
    \]
    where:
    \begin{itemize}
        \item \( \text{deg}(v) \): Degree of node \( v \), which is the number of direct connections (or edges) it has.
        \item \( N \): Total number of nodes in the graph.
    \end{itemize}

    \item \textbf{Betweenness Centrality}: Quantifies the number of times a node acts as a bridge along the shortest path between two other nodes.
    \[
    \text{Betweenness Centrality}(v) = \sum_{s \neq v \neq t} \frac{\sigma_{st}(v)}{\sigma_{st}}
    \]
    where:
    \begin{itemize}
        \item \( s \) and \( t \): Nodes in the graph, distinct from \( v \).
        \item \( \sigma_{st} \): Total number of shortest paths from node \( s \) to node \( t \).
        \item \( \sigma_{st}(v) \): Number of shortest paths from \( s \) to \( t \) that pass through node \( v \).
    \end{itemize}

    \item \textbf{Closeness Centrality}: Represents how close a node is to all other nodes in the network, based on the shortest paths.
    \[
    \text{Closeness Centrality}(v) = \frac{1}{\sum_{u \neq v} d(u, v)}
    \]
    where:
    \begin{itemize}
        \item \( u \): Another node in the graph, distinct from \( v \).
        \item \( d(u, v) \): Shortest path distance between node \( u \) and node \( v \).
    \end{itemize}

    \item \textbf{Eigenvector Centrality}: Measures influence by considering a node’s connections to other influential nodes.
    \[
    \text{Eigenvector Centrality}(v) = \frac{1}{\lambda} \sum_{u \in \mathcal{N}(v)} A_{vu} x_u
    \]
    where:
    \begin{itemize}
        \item \( \lambda \): Largest eigenvalue of the adjacency matrix \( A \).
        \item \( \mathcal{N}(v) \): Set of neighbors (directly connected nodes) of \( v \).
        \item \( A_{vu} \): Entry in the adjacency matrix representing the edge between nodes \( v \) and \( u \) (1 if there’s an edge, 0 if not).
        \item \( x_u \): Eigenvector centrality of node \( u \).
    \end{itemize}

    \item \textbf{Harmonic Centrality}: Sum of the reciprocal shortest path distances from all other nodes to a given node.
    \[
    \text{Harmonic Centrality}(v) = \sum_{u \neq v} \frac{1}{d(u, v)}
    \]
    where:
    \begin{itemize}
        \item \( u \): Another node in the graph, distinct from \( v \).
        \item \( d(u, v) \): Shortest path distance from node \( u \) to node \( v \).
    \end{itemize}

    \item \textbf{Load Centrality}: Fraction of all shortest paths that pass through a node, representing its role in traffic flow.
    \[
    \text{Load Centrality}(v) = \frac{\text{Number of shortest paths through } v}{\text{Total shortest paths in the network}}
    \]

    \item \textbf{PageRank}: Assigns a probability score to each node based on its link structure, indicating the likelihood of arriving at a node by randomly following edges.
    \[
    \text{PageRank}(v) = \frac{1 - d}{N} + d \sum_{u \in \mathcal{N}(v)} \frac{\text{PageRank}(u)}{\text{outdeg}(u)}
    \]
    where:
    \begin{itemize}
        \item \( d \): Damping factor (typically around 0.85), representing the probability of following an edge.
        \item \( N \): Total number of nodes in the graph.
        \item \( \mathcal{N}(v) \): Set of neighbors of node \( v \) (nodes pointing to \( v \)).
        \item \( \text{outdeg}(u) \): Out-degree of node \( u \), or the number of edges going out from \( u \).
    \end{itemize}

    \item \textbf{Clustering Coefficient}: Measures the degree to which nodes in a network tend to cluster together, based on the number of triangles around each node.
    \[
    \text{Clustering Coefficient}(v) = \frac{2T(v)}{\text{deg}(v)(\text{deg}(v) - 1)}
    \]
    where:
    \begin{itemize}
        \item \( T(v) \): Number of triangles through node \( v \), or the number of closed triplets (three nodes all connected).
        \item \( \text{deg}(v) \): Degree of node \( v \), which is the number of direct connections it has.
    \end{itemize}
\end{itemize}

\subsection{Datasets}
\label{Dataset_details}

\subsubsection{SNAPSHOT Data}
The SNAPSHOT dataset is a longitudinal multimodal dataset collected over five years from 250 participants (158 female, 92 male) with an average age of 21 years. It includes data from wrist-worn sensors (skin conductance, temperature, and acceleration), phone metadata (calls, SMS, screen usage, and location), and daily surveys covering sleep, exercise, and social interactions. Each participant contributed data for a time period of up to 110 days, depending on their cohort. Details of data processing, feature engineering and data statistics are available in \cite{sleepnet}.

\subsubsection{Friends and Family study Dataset}
The Friends and Family dataset offers a comprehensive, multi-modal collection of data focusing on social behavior and decision-making within a residential community adjacent to a major research university in North America\cite{socialFMRI}. The study was conducted in two phases. \textbf{Phase I}, starting in Spring 2010, included 55 participants, while \textbf{Phase II}, launched in Fall 2010, expanded the study to include 130 participants across approximately 64 families. Data collection was conducted over a 15-month period using Android-based mobile phones equipped with various sensors to track the subjects’ activities and interactions.

The dataset contains detailed logs of multiple types of sensor data and survey responses, each providing insights into participants’ activities, interactions, and social networks. 

\textbf{1. Sensor Data}
\begin{itemize}
    \item \textbf{Bluetooth Proximity}: Records the nearby Bluetooth devices every 5 minutes, including participant IDs and anonymized MAC addresses.
    \item \textbf{Call Logs}: Includes call details such as participant IDs, call times, types (incoming, outgoing, missed), and hashed phone numbers.
    \item \textbf{SMS Logs}: Tracks incoming and outgoing text messages with participant IDs and hashed phone numbers.
    \item \textbf{Location Data}: Captured every 30 minutes, providing time-stamped location information with confidence offsets for accuracy.
    \item \textbf{App Data}: Logs the installed apps (scanned every 10 minutes to 3 hours), running apps (scanned every 30 seconds), app package names, and installation status.
    \item \textbf{Accelerometer Data}: Captures phone accelerations at various intervals, indicating physical activity, with additional information about exercise activities.
    \item \textbf{Battery Status}: Logs battery levels, charging states, and power sources to understand usage patterns.
\end{itemize}

\subsection*{2. Surveys}
\begin{itemize}
    \item \textbf{Monthly Surveys}: Gathered from 2010 to 2011, these include self-reported data on demographics, personality (e.g., Big-5 survey), social dynamics, and financial perceptions.
    \item \textbf{Weekly Surveys}: Collected information on aspects such as daily sleep quality, daily mood, and social activities (dining, TV watching, babysitting).
    \item \textbf{Social Network Data}: Surveys identified friendship ties, with the dataset capturing self-reported closeness between participants and specific details about couples and their interactions.
\end{itemize}

\subsubsection{Engineering Call/SMS Features}

This section describes the process used to preprocess and engineer features from both call and SMS data, focusing on extracting daily communication patterns and incorporating relevant interactions between study participants. The call and SMS datasets include information from two participants in each interaction, labeled as \textit{participantID.A} and \textit{participantID.B}. This feature engineering process aims to handle missing values, reverse roles between participants where applicable, and calculate daily and time-windowed communication features for further analysis.

The primary steps in the feature engineering process are as follows:

\begin{itemize}
    \item \textbf{Role Reversal Between Participants:} In both call and SMS datasets, when \textit{participantID.B} is not missing, we generate a new entry where \textit{participantID.B} is treated as the main participant. In this new record, the roles of \textit{participantID.A} and \textit{participantID.B} are swapped, effectively capturing the interaction from the perspective of \textit{participantID.B}. This is especially useful for understanding reciprocal communication patterns. The type of interaction (e.g., incoming or outgoing) is also reversed in this process. For example, if a record for \textit{participantID.A} is labeled as \textit{incoming}, the corresponding entry for \textit{participantID.B} is set to \textit{outgoing}. This applies to both call and SMS data.
    
    \item \textbf{Handling Call Types:} In the call dataset, we handled three types of calls: \textit{incoming}, \textit{outgoing}, and \textit{missed}. For records where \textit{participantID.B} initiated a \textit{missed} call, we treated the call type as \textit{outgoing} for \textit{participantID.B} when reversing the roles between participants. Additionally, any missing duration values were filled with zero, as missed calls inherently have zero duration.
    
    \item \textbf{Time-Window Categorization:} Each communication event (both call and SMS) is associated with a timestamp, which we used to classify the interactions into four 6-hour time windows within each day: 00:00–06:00, 06:00–12:00, 12:00–18:00, and 18:00–00:00. This allowed us to capture the distribution of communication events across different periods of the day, which can provide insights into daily behavioral patterns.
    
    \item \textbf{Feature Extraction:} For each participant, we aggregated the number of \textit{incoming}, \textit{outgoing}, and \textit{missed} calls and SMS messages on a daily basis, as well as within the predefined time windows. The total communication duration for calls was also computed for each time window and for the entire day. Specifically, the following features were extracted:
    \begin{itemize}
        \item Total number of calls (both incoming and outgoing) per day and within each time window.
        \item Total number of SMS messages (both incoming and outgoing) per day and within each time window.
        \item Total duration of calls per day and within each time window.
        \item Total number of missed calls per day and within each time window.
    \end{itemize}
    
    \item \textbf{Daily and Time-Window Aggregation:} After reversing participant roles and classifying events into time windows, we aggregated the data to calculate daily features for each participant. For calls, we computed the total number of incoming, outgoing, and missed calls and the total call duration. We computed the total number of incoming and outgoing messages for SMS data. These features were computed for the entire day and each of the four time windows. Additionally, we calculated the total number of communication events (calls and SMS) for each participant daily, capturing overall communication activity.
    
    \item \textbf{Merging Call and SMS Data:} Finally, we merged the daily and time-window features from the call and SMS datasets to create a unified dataset that includes both types of communication. This provided a comprehensive view of each participant’s communication patterns on a daily basis, integrating both call and SMS interactions.
\end{itemize}

After merging the call and SMS datasets with the original dataset, some entries had missing values for the call and SMS features, indicating that no communication (calls or SMS) occurred for certain users on specific days. To handle these missing values in the analysis, we filled all missing values with zero. This transformation implies that on days when a participant has no recorded communication activity, their call and SMS feature values (e.g., total number of calls, total call duration, total SMS messages) are set to zero.

\subsection{Ground Truth Collection and Processing}
\subsubsection{Mood}
To capture users' predominant moods during the day, participants were asked to select from five possible mood options: \textit{Happy or content}, \textit{Sad or depressed}, \textit{Calm or peaceful}, \textit{Stressed or anxious}, and \textit{Angry or frustrated}. Each user could select one or more mood states that best described their feelings during the day. 

To process these responses, the mood selections were encoded in a binary format where each mood option was represented by a unique bit position. For each user response, a binary string was generated by assigning `1` if a mood was selected and `0` if it was not. This binary string was then converted to a unique integer between 1 and 32, with `1` representing the selection of \textit{Happy or content} only, `2` representing \textit{Sad or depressed} only, and so on. The final integer was stored as a \textit{mood label}, representing each response's specific combination of mood selections.
\begin{figure}[!]
    \centering
    \includegraphics[width=0.99\textwidth]{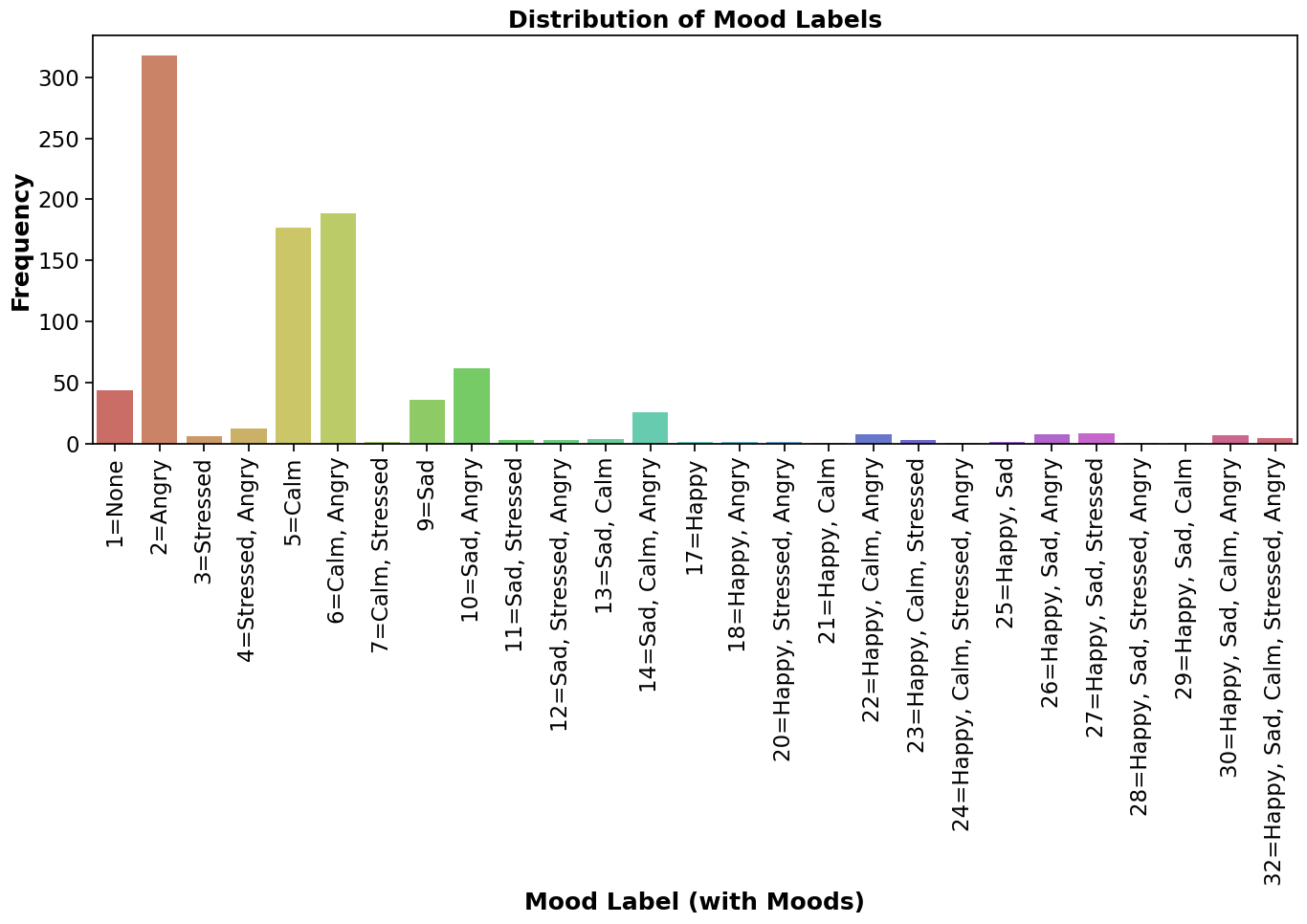}
    \caption{Mood label distribution: the x-axis shows mood combinations (e.g., 2=Angry, 4=Stressed+Angry), and the y-axis shows their frequency.}
    \label{fig:mood label distribution}
\end{figure}
To visualize the distribution of mood combinations across participants, a bar plot was generated where each label on the x-axis represented a unique combination of selected moods, and the y-axis showed the frequency of occurrence. The mood labels were presented in a compact format, e.g., \textit{1=Happy}, \textit{3=Happy+Sad}, for ease of interpretation. This method enabled efficient encoding and visualization of multiple mood selections for each participant. As observed from the plot, the data is highly imbalanced. Therefore, instead of using multi-mood labels, we use binary labels and evaluate the models for different moods independently.

\subsection{Sleep Distribution and Classification Strategy}

The dataset contains self-reported sleep duration data collected through surveys. Participants answered the following question:  
\textit{"To the best of your recollection, how many hours did you sleep on each of the following nights?"}  
Respondents selected from the following options: \textless 5, 5, 6, 7, 8, 9, \textgreater 9.

\begin{figure}[h!]
    \centering
    \includegraphics[width=0.8\textwidth]{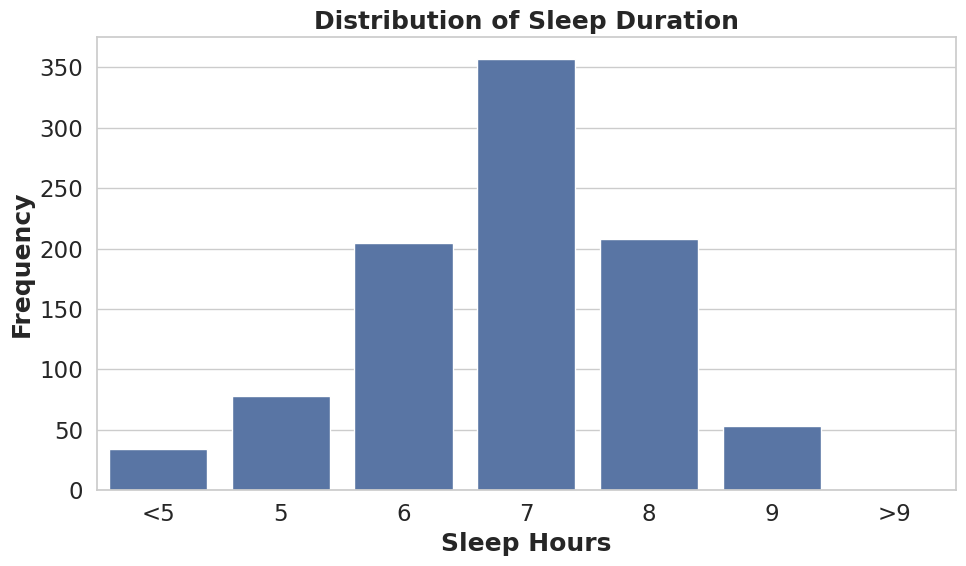}
    \caption{Distribution of reported sleep hours among participants. The x-axis represents hours of sleep, and the y-axis represents the frequency of participants in each sleep category.}
    \label{fig:sleep_distribution}
\end{figure}

After cleaning the data, the distribution of sleep hours was visualized, revealing distinct patterns in participants' sleep behavior. As shown in Figure \ref{fig:sleep_distribution}, the responses formed three key groupings: a group that reported sleeping 6 hours or less, a significant number of participants who reported sleeping exactly 7 hours, and another group that reported sleeping more than 7 hours.

To facilitate further analysis, we categorized the sleep data into two distinct classes:
\begin{itemize}
    \item \textbf{Class 0}: Participants who reported sleeping \textbf{6 hours or less}.
   
    \item \textbf{Class 1}: Participants who reported sleeping \textbf{more than 6 hours}.
\end{itemize}

%These cutoffs provided a balanced dataset with similar numbers in each class: 317 participants in class 0, 357 participants in class 1, and  261 participants in class 2. This classification enables a structured analysis of sleep patterns and their possible effects on participant well-being. The chosen cutoffs create a balanced dataset, which is particularly useful for the robust evaluation of machine learning models. Research suggests that both insufficient and excessive sleep can have varying impacts on health, affecting cognitive performance, emotional stability, and physical well-being. While around seven hours of sleep is often considered ideal, the optimal duration can vary depending on individual needs and lifestyle factors. Both very short and long sleep durations have been linked to potential health risks, but the exact thresholds may differ across studies.

%We also create binary labels by distinguishing between sleep labels  less than 7 hours (Class 0) and those equal to 7 hours or more (Class 1).

\section{Additional Results}

\subsection{Comparative Analysis and Network Topology}
\begin{figure}
    \centering
    \includegraphics[width=0.9\linewidth]{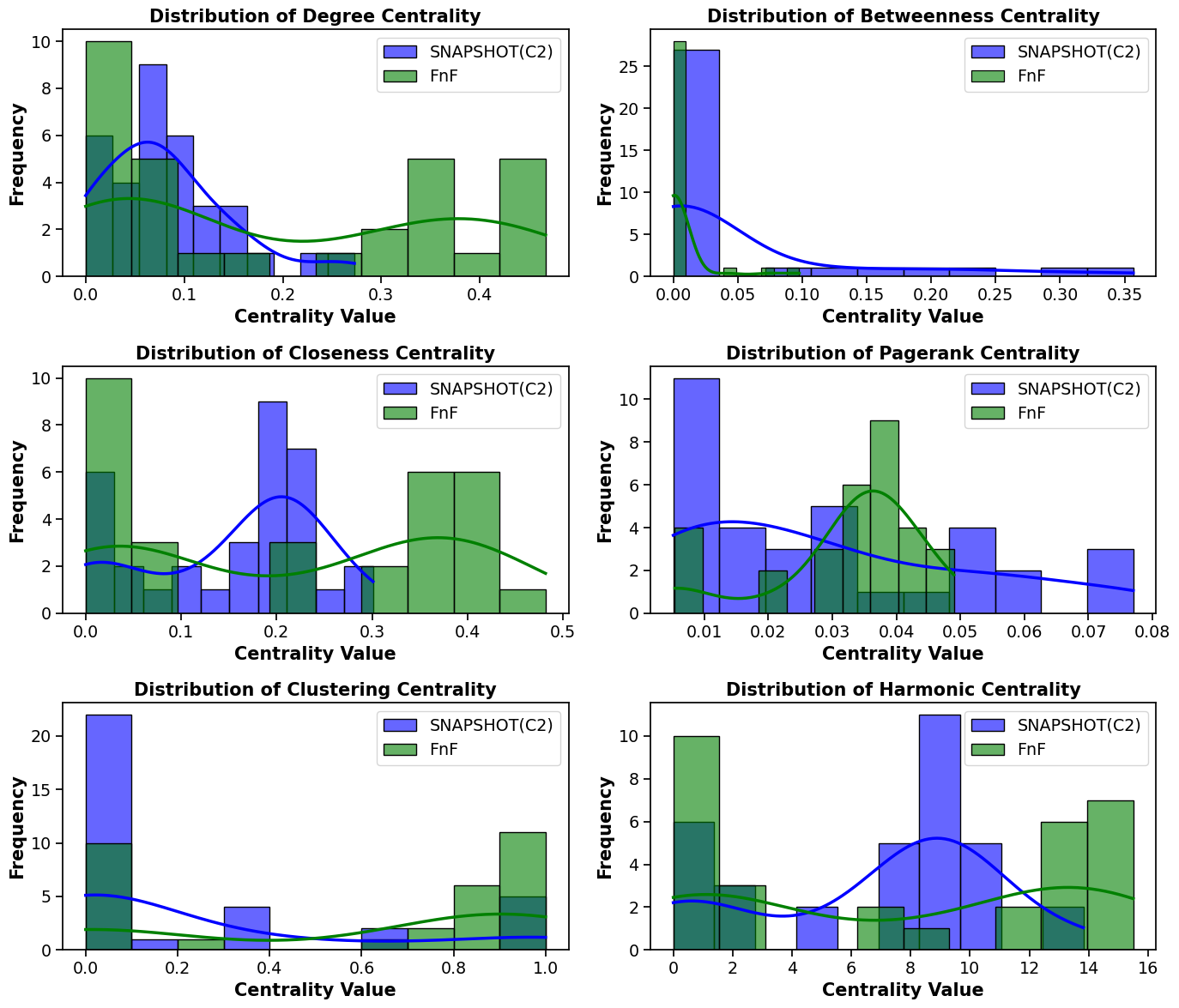}
    \caption{Variation in Network Topology in Two Datasets}
    \label{fig:topology}
\end{figure}

\end{document}

\maketitle

\section{Results for FnF Dataset}
In this section, we present detailed results showing how each active learning (AL) strategy performs across different node types—both overall and as performance evolves over time through successive AL iterations, as new nodes are queried and the model is updated with additional labeled data. Additionally, we analyze the user burden and diversity associated with each strategy for the sleep classification task in the FnF dataset.
\subsection{Average Performance Comparison Across Node Types}

\begin{table}[b]
\centering
\resizebox{\columnwidth}{!}{%
\begin{tabular}{|p{2.8cm}|p{2.5cm}|p{2.5cm}|p{2.5cm}|p{2.5cm}|}
\hline
\textbf{AL Strategy} & \textbf{Test Set (Same Day)} & \textbf{Unqueried Nodes (Same Day)} & \textbf{Unqueried Nodes (Next Day)} & \textbf{Train Nodes (Next Day)} \\ \hline
random\_sample & 0.39 (0.03) & 0.42 (0.02) & 0.41 (0.02) & 0.54 (0.04) \\ \hline
no AL (baseline) & 0.45 (0.05) & 0.41 (0.03) & 0.40 (0.02) & N/A \\ \hline
graphpart & 0.36 (0.04) & \textbf{0.48 (0.01)} & 0.43 (0.01) & 0.41 (0.03) \\ \hline
graphpartfar & 0.66 (0.00) & 0.37 (0.00) & 0.42 (0.02) & \textbf{0.97 (0.01)} \\ \hline
age & \textbf{0.69 (0.03)} & 0.43 (0.01) & 0.42 (0.01) & 0.58 (0.02) \\ \hline
coreset & 0.41 (0.03) & 0.44 (0.01) & 0.43 (0.02) & 0.55 (0.04) \\ \hline
featProp & 0.45 (0.02) & 0.44 (0.01) & 0.38 (0.01) & 0.46 (0.03) \\ \hline
density & \textbf{0.68 (0.03)} & 0.44 (0.01) & 0.42 (0.01) & 0.58 (0.03) \\ \hline
uncertainty entropy & 0.42 (0.05) & 0.47 (0.02) & 0.44 (0.02) & 0.52 (0.04) \\ \hline
uncertainty least confidence & 0.41 (0.05) & 0.46 (0.02) & 0.43 (0.02) & 0.53 (0.04) \\ \hline
uncertainty margin & 0.45 (0.03) & 0.43 (0.03) & \textbf{0.45 (0.03)} & 0.47 (0.04) \\ \hline
pagerank & 0.44 (0.03) & 0.45 (0.02) & 0.44 (0.02) & 0.62 (0.04) \\ \hline
degree & 0.44 (0.03) & 0.45 (0.02) & 0.44 (0.02) & 0.62 (0.04) \\ \hline
\end{tabular}%
}
\caption{CPI Comparison for Active Learning Strategies (Sleep Label, L=6, k=5)}
\label{tab:cpi_comparison_sleep_label}
\end{table}

\begin{table}[]
\centering
\resizebox{\columnwidth}{!}{%
\begin{tabular}{|p{2.8cm}|p{2.5cm}|p{2.5cm}|p{2.5cm}|p{2.5cm}|}
\hline
\textbf{AL Strategy} & \textbf{Test Set (Same Day)} & \textbf{Unqueried Nodes (Same Day)} & \textbf{Unqueried Nodes (Next Day)} & \textbf{Train Nodes (Next Day)} \\ \hline

random\_sample & 0.41 (0.18) & 0.44 (0.12) & 0.42 (0.12) & 0.56 (0.25) \\ \hline
no AL (baseline) & 0.47 (0.18) & 0.44 (0.10) & 0.43 (0.11) & N/A \\ \hline
graphpart & 0.38 (0.20) & \textbf{0.51 (0.12)} & 0.44 (0.13) & 0.45 (0.35) \\ \hline
graphpartfar & \textbf{0.70 (0.28)} & 0.39 (0.04) & 0.39 (0.04) & \textbf{0.98 (0.01)} \\ \hline
age & 0.68 (0.25) & 0.44 (0.09) & 0.44 (0.10) & 0.60 (0.29) \\ \hline
coreset & 0.42 (0.20) & 0.46 (0.11) & 0.45 (0.12) & 0.57 (0.26) \\ \hline
featProp & 0.47 (0.19) & 0.46 (0.11) & 0.40 (0.13) & 0.48 (0.24) \\ \hline
density & 0.69 (0.26) & 0.45 (0.09) & 0.43 (0.09) & 0.60 (0.30) \\ \hline
uncertainty entropy & 0.43 (0.19) & 0.48 (0.13) & 0.45 (0.14) & 0.55 (0.25) \\ \hline
uncertainty least confidence & 0.42 (0.19) & 0.48 (0.13) & 0.45 (0.14) & 0.55 (0.25) \\ \hline
uncertainty margin & 0.46 (0.22) & 0.45 (0.12) & 0.45 (0.13) & 0.49 (0.24) \\ \hline
pagerank & 0.46 (0.22) & 0.46 (0.13) & \textbf{0.48 (0.12)} & 0.64 (0.31) \\ \hline
degree & 0.46 (0.22) & 0.46 (0.13) & \textbf{0.48 (0.12)} & 0.64 (0.31) \\ \hline
\end{tabular}%
}
\caption{F1 Macro Comparison for Active Learning Strategies (Sleep Label, L=6, k=5)}
\label{tab:f1_macro_comparison_sleep}
\end{table}

\begin{table}[]
\centering
\resizebox{\columnwidth}{!}{%
\begin{tabular}{|p{2.8cm}|p{2.5cm}|p{2.5cm}|p{2.5cm}|p{2.5cm}|}
\hline
\textbf{AL Strategy} & \textbf{Test Set (Same Day)} & \textbf{Unqueried Nodes (Same Day)} & \textbf{Unqueried Nodes (Next Day)} & \textbf{Train Nodes (Next Day)} \\ \hline
random\_sample & 0.56 (0.22) & 0.57 (0.13) & 0.55 (0.14) & 0.66 (0.21) \\ \hline
no AL (baseline) & 0.70 (0.17) & 0.51 (0.12) & 0.52 (0.14) & N/A \\ \hline
graphpart & 0.49 (0.25) & 0.55 (0.12) & 0.53 (0.14) & 0.54 (0.33) \\ \hline
graphpartfar & \textbf{0.89 (0.14)} & \textbf{0.64 (0.10)} & \textbf{0.64 (0.10)} & \textbf{0.97 (0.01)} \\ \hline
age & 0.86 (0.12) & 0.62 (0.09) & 0.63 (0.10) & 0.77 (0.21) \\ \hline
coreset & 0.58 (0.24) & 0.57 (0.13) & 0.56 (0.15) & 0.66 (0.23) \\ \hline
featProp & 0.66 (0.21) & 0.56 (0.14) & 0.50 (0.18) & 0.61 (0.23) \\ \hline
density & 0.85 (0.14) & 0.63 (0.09) & 0.62 (0.12) & 0.77 (0.23) \\ \hline
uncertainty entropy & 0.61 (0.22) & 0.57 (0.15) & 0.55 (0.17) & 0.65 (0.21) \\ \hline
uncertainty least confidence & 0.60 (0.22) & 0.57 (0.15) & 0.54 (0.16) & 0.65 (0.22) \\ \hline
uncertainty margin & 0.62 (0.22) & 0.56 (0.15) & 0.58 (0.15) & 0.63 (0.23) \\ \hline
pagerank & 0.62 (0.23) & 0.55 (0.13) & 0.55 (0.13) & 0.78 (0.25) \\ \hline
degree & 0.62 (0.23) & 0.55 (0.13) & 0.55 (0.13) & 0.78 (0.25) \\ \hline
\end{tabular}%
}
\caption{Accuracy Comparison for Active Learning Strategies (Sleep Label, L=6, k=5)}
\label{tab:accuracy_comparison_sleep}
\end{table}

The results for CPI evaluation for L=6 and k=5 for the friendship social network on mood prediction are presented in Table \ref{tab:cpi_comparison_sleep_label}. The choice of \( L = 6 \) addresses the cold start problem in AL, where too few initial labeled samples (e.g., \( L = 1,2,3,4,5 \)) limit the model’s capacity to learn meaningful patterns. Setting \( k = 5 \) ensures sufficient samples per round for effective retraining, as very small \( k \) values make it challenging to distinguish between different AL strategies, reducing their impact on model performance.
The Holdout Test Set and Unqueried Nodes metrics across these tables provide key insights into the robustness of each AL strategy. In particular, Table \ref{tab:cpi_comparison_sleep_label} shows that graphpartfar excels in both Train Nodes and Unqueried Nodes (Next Day), demonstrating its effectiveness in learning from training data and extending this knowledge to data not previously seen. Age consistently performs well in generalization, with strong scores across all unqueried and test categories. These findings indicate that graphpartfar and age are optimal choices for applications that require reliable training outcomes and strong adaptability to unseen data.

The F1 Macro scores in Table \ref{tab:f1_macro_comparison_sleep} highlight the strengths of various Active Learning (AL) strategies in predicting across different data categories. Notably, graphpartfar achieves the highest scores for Train Nodes (Next Day), indicating its ability to capture essential patterns in training data and make reliable future predictions. Density also stands out in the Holdout Test Set, suggesting its effectiveness in generalizing to entirely unseen data, which is crucial for assessing model robustness. Additionally, age, degree, and PageRank perform well in unqueried node categories, showing its capability to handle unseen data on the same day and predict accurately across time steps. These results suggest that graphpartfar, density, and age are highly effective for applications that require both training accuracy and generalization.

The accuracy results in Table \ref{tab:accuracy_comparison_sleep} reinforce the value of graphpartfar and age as top-performing strategies, particularly on the Holdout Test Set and Unqueried Nodes(Same Day). Graphpartfar’s high accuracy scores across training and testing data categories demonstrate its stability in predictions and robust generalization capabilities, making it a reliable choice for handling dynamic and unseen data. Density and age also show promising accuracy on unqueried nodes, indicating that these strategies can adapt effectively to data that was excluded from training, further supporting their utility in applications requiring adaptability and generalization. Overall, graphpartfar leads in accuracy, followed closely by density and age, both of which are effective alternatives in contexts where generalization is essential.

The statistical analysis in Table~\ref{tab:stat_results_node_types_Sleep} indicates significant differences across strategies, confirming that each strategy produces distinct performance outcomes for accuracy, F1 Macro, and CPI metrics.

\subsection{Daily Variation Analysis of Active Learning Strategies by Node
Types and Metrics}

\begin{figure}[htbp!]
    \centering
    \includegraphics[width=\textwidth,height= \textheight]{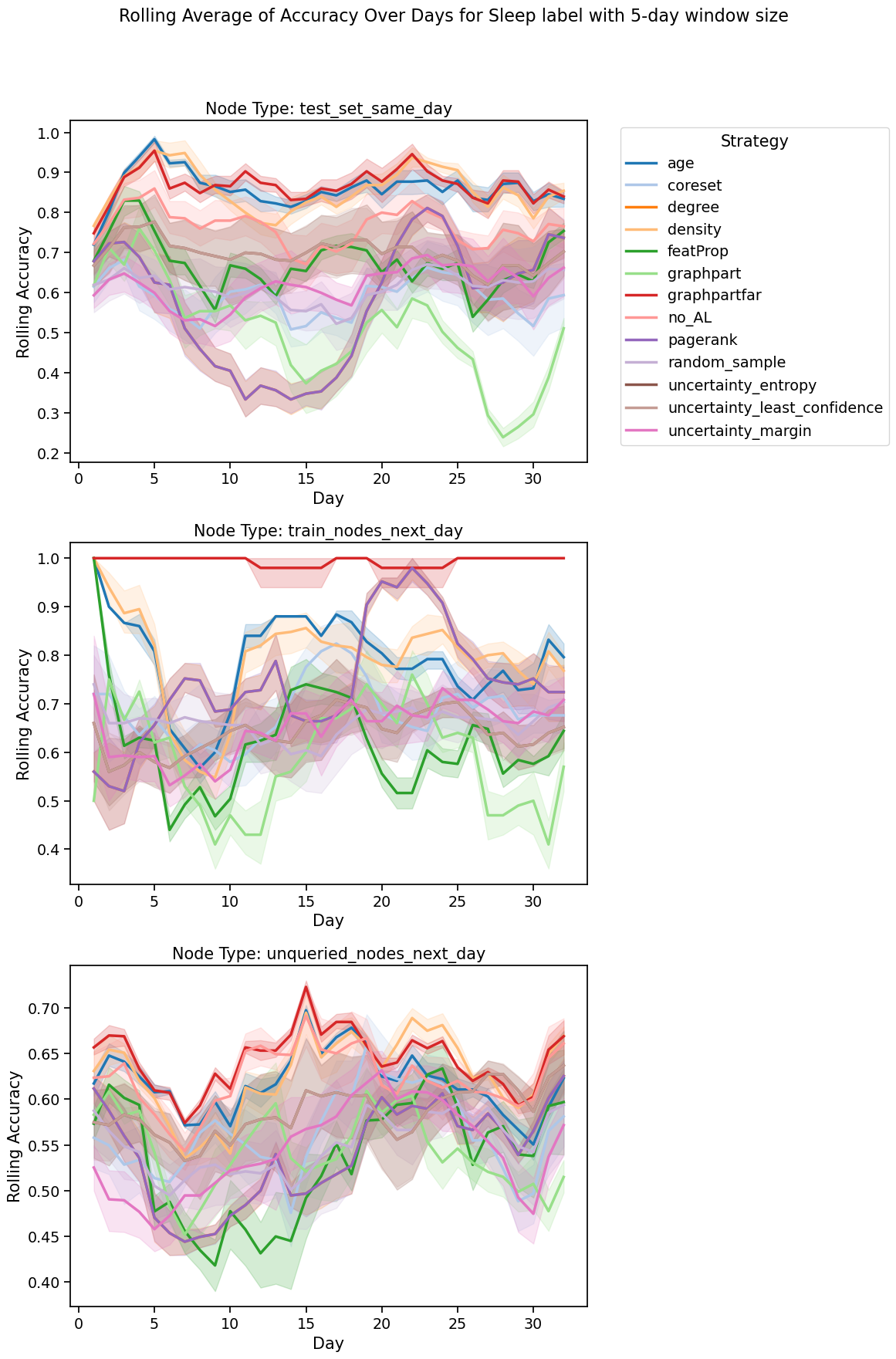}
    \caption{Rolling Average and standard deviation of Accuracy Over Days for Sleep Label with 5-day Window Size for L=6 and k=8}
    \label{fig:rolling_sleep}
\end{figure}

The plot in Figure~\ref{fig:rolling_sleep} illustrates the rolling average of accuracy over days for the sleep label, separated by node types ("test set same day," "train nodes next day," and "unqueried nodes next day") with a 5-day rolling window.

The initial spike in performance occurs because, during the first six days, the model is trained on all available data, providing it with a comprehensive starting point and resulting in high initial accuracy. After this phase, active learning (AL) strategies are implemented, where each strategy selectively queries only 8 nodes out of the 31 available, shifting to a more limited and targeted approach to learning. This selective querying causes an initial drop in performance as the model adjusts to using smaller, more focused data samples.

As time progresses, accuracy begins to improve as the model learns to generalize from the limited but informative samples queried by each strategy. Some AL strategies like graphpartfar, density, and age exhibit faster learning and recovery in accuracy than others, indicating differences in their effectiveness.

Additionally, we observe periodic drops in accuracy after every few days, which can be attributed to changes in behavioral patterns or shifts in data trends. These fluctuations suggest that the model encounters new behaviors in the data, causing temporary declines in accuracy. However, as the model acquires more training data through additional queries, it adapts and recovers, learning to adjust to these evolving patterns over time. 

\subsection{Evaluating Diversity \& Sampling Burden across Multiple AL Strategies }

\begin{figure}
    \centering
    \includegraphics[width=1.0\linewidth]{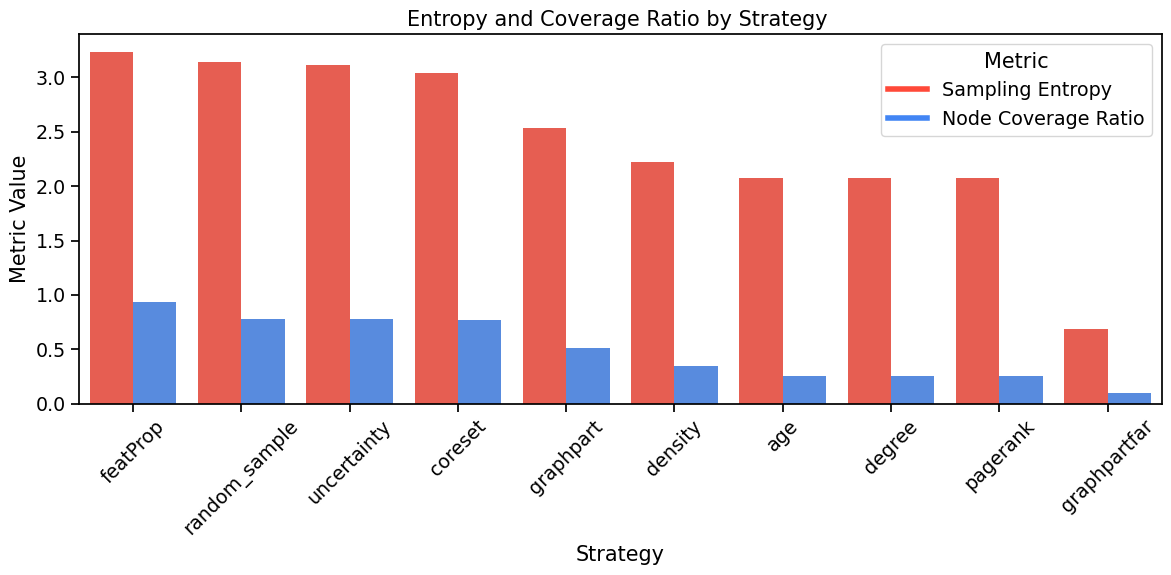}
    \caption{Sampling Entropy and Node Coverage Ratio for various Active Learning Strategies in Sleep Classification}
    \label{fig:entropy_coverage_sleep}
\end{figure}

The results in the Figure. \ref{fig:entropy_coverage_sleep} for sleep classification offers insights into the performance of various Active Learning (AL) strategies in terms of Sampling Entropy and Node Coverage Ratio. Strategies such as featProp, random sampling, and uncertainty demonstrate high Sampling Entropy, indicating a balanced selection process across the graph, which helps to prevent over-sampling specific nodes and ensures a broad, representative sample of data points. This approach is valuable for capturing diverse data patterns across the network, which is essential in sleep classification to account for variability in individual sleep behaviors. 

On the other hand, strategies like pagerank and graphpartfar exhibit lower entropy, suggesting a more targeted focus on certain influential nodes, which may benefit scenarios requiring prioritized insights from key individuals but could limit the overall representativeness of the data. In terms of Node Coverage Ratio, random sampling and featProp achieve higher coverage, capturing information from a wide array of nodes. This comprehensive coverage supports robust sleep classification by gathering diverse data across the network, while strategies like pagerank and graphpartfar, with lower coverage, risk overlooking less central nodes, potentially impacting the model’s ability to generalize sleep patterns across all users.

\begin{figure}
    \centering
    \includegraphics[width=1.0\linewidth]{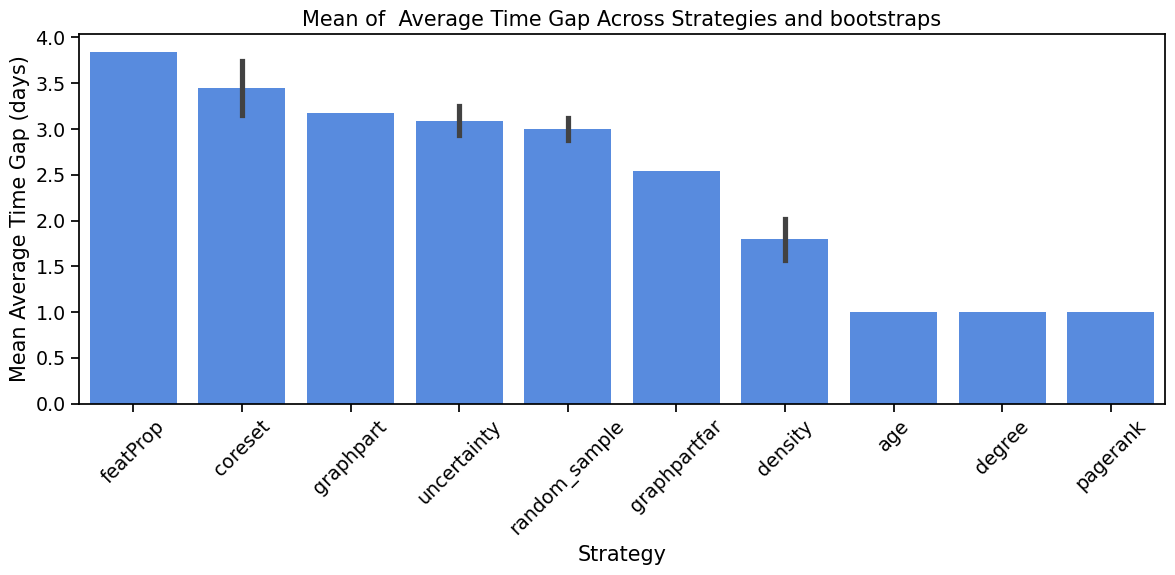}
    \caption{Mean Average Time Gap across Active Learning Strategies for Sleep Classification}
    \label{fig:avg_time_gap_sleep}
\end{figure}

The Average Time Gap plot for sleep classification in Figure \ref{fig:avg_time_gap_sleep} illustrates the mean interval, in days, between consecutive samples for each node, providing insights into how effectively each Active Learning strategy distributes its sampling over time. Strategies like featProp and coreset show the longest average time gaps, suggesting that they space out their sampling, which could reduce the burden on users by minimizing frequent queries. This approach may be useful for capturing long-term changes in sleep patterns. In contrast, strategies like pagerank, degree, and age have shorter time gaps, indicating more frequent sampling of specific nodes. 

Comparing the Average Time Gap plot with Sampling Entropy and Node Coverage Ratio provides valuable insights into the balance each Active Learning strategy strikes between temporal sampling distribution and spatial node coverage.

Strategies such as featProp and coreset, which demonstrated high entropy and coverage, also exhibit longer average time gaps in sampling. This suggests that these strategies not only cover a diverse range of nodes but also allow for spaced-out sampling intervals, which can be advantageous in real-world applications by reducing user burden and avoiding excessive sampling of the same nodes. This balance makes featProp and coreset ideal for applications that prioritize broad network coverage with minimized user engagement over time.

In contrast, strategies like pagerank, degree, and age show both lower entropy and shorter average time gaps. These strategies tend to concentrate on a smaller subset of highly connected or influential nodes, sampling them more frequently. While this focused approach may capture critical changes within key nodes, it risks overlooking the broader network and can lead to over-sampling certain users, potentially causing user fatigue.

\begin{figure}
    \centering
    \includegraphics[width=1.0\linewidth]{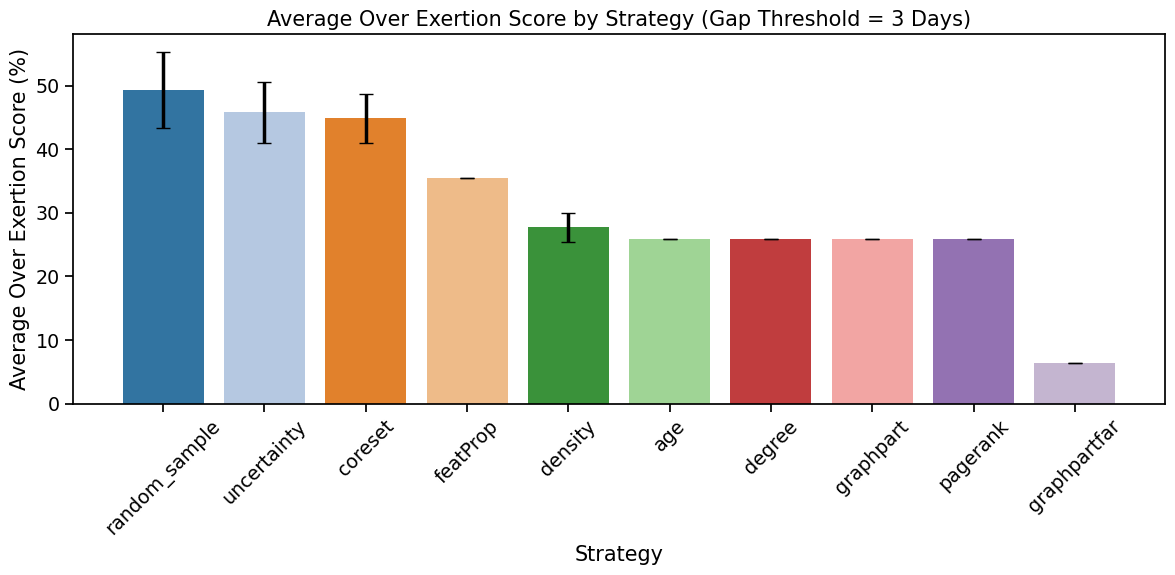}
    \caption{Average Over Exertion Score by strategy at a fixed 3-day gap threshold for Sleep Classification}
    \label{fig:over_exer_sleep}
\end{figure}

The Average Over Exertion Score for sleep classification (Gap Threshold = 3 Days) in the Figure \ref{fig:over_exer_sleep} highlights how frequently each Active Learning strategy samples nodes within a short interval, potentially causing user fatigue or resource depletion in practical applications. Strategies such as random sampling, coreset, and uncertainty exhibit high over-exertion scores, indicating they frequently re-sample nodes within the 3-day gap. 

In contrast, strategies like graphpart and graphpartfar show much lower over-exertion scores, demonstrating a more restrained sampling approach that reduces the risk of exhausting individual nodes within the 3-day period. These strategies may be preferable in scenarios where balanced sampling and lower re-sampling frequency are critical for user or resource retention. 

\begin{figure}
    \centering
    \includegraphics[width=1.0\linewidth]{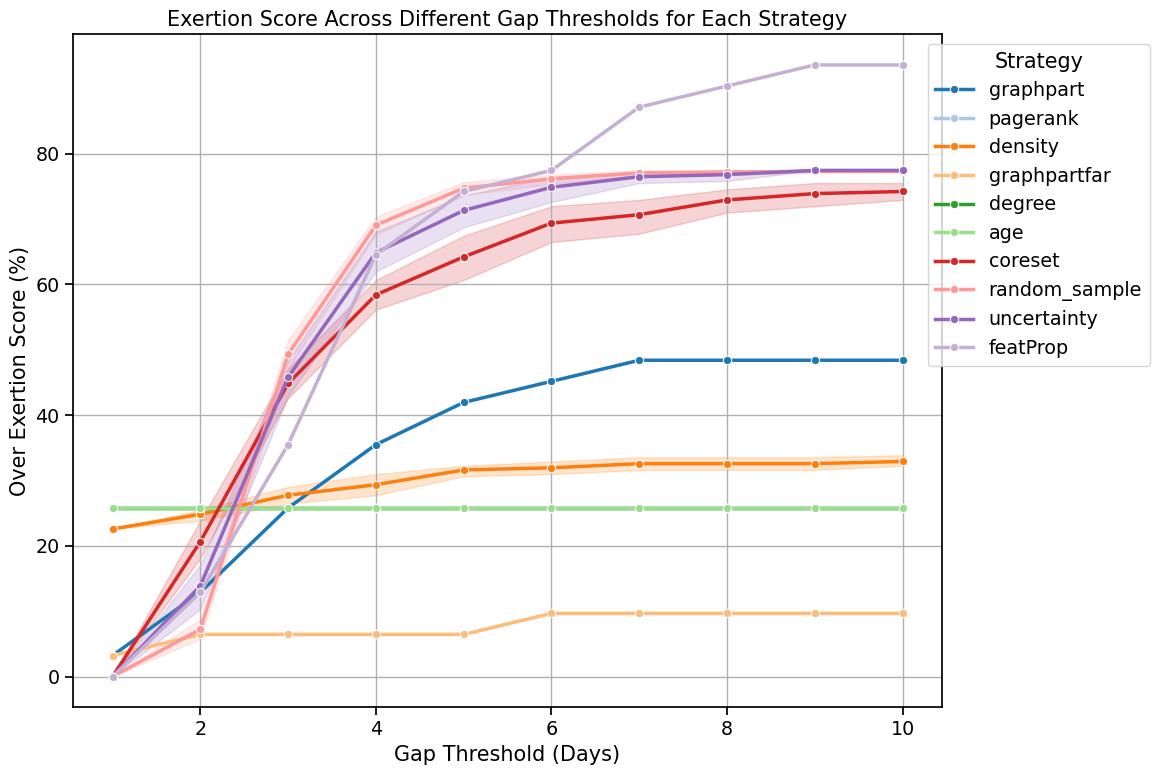}
    \caption{Variation of Over Exertion Score across different gap thresholds for Sleep Classification}
    \label{fig:over_exer_sleep_varygap}
\end{figure}

The Over-exertion Score across different gap thresholds in Figure \ref{fig:over_exer_sleep_varygap} further illustrates how each strategy's sampling pattern adapts as the allowable interval between queries increases. While most strategies show higher over-exertion scores at longer gap thresholds, some, such as graphpart and graphpartfar, maintain low exertion levels even as the gap increases, indicating a consistent, restrained approach. Conversely, strategies like featProp, coreset, and uncertainty significantly increase their over-exertion scores at longer gaps, suggesting a focus on frequent updates of specific nodes. This trade-off highlights the importance of selecting an AL strategy that matches the system's priorities, whether that’s frequent data collection for high-resolution monitoring or balanced load distribution to ensure sustainable usage.

\subsection{Centrality Preferences and Exertion Patterns
Across Sampling Strategies}

\begin{figure}
    \centering
    \includegraphics[width=1.0\linewidth]{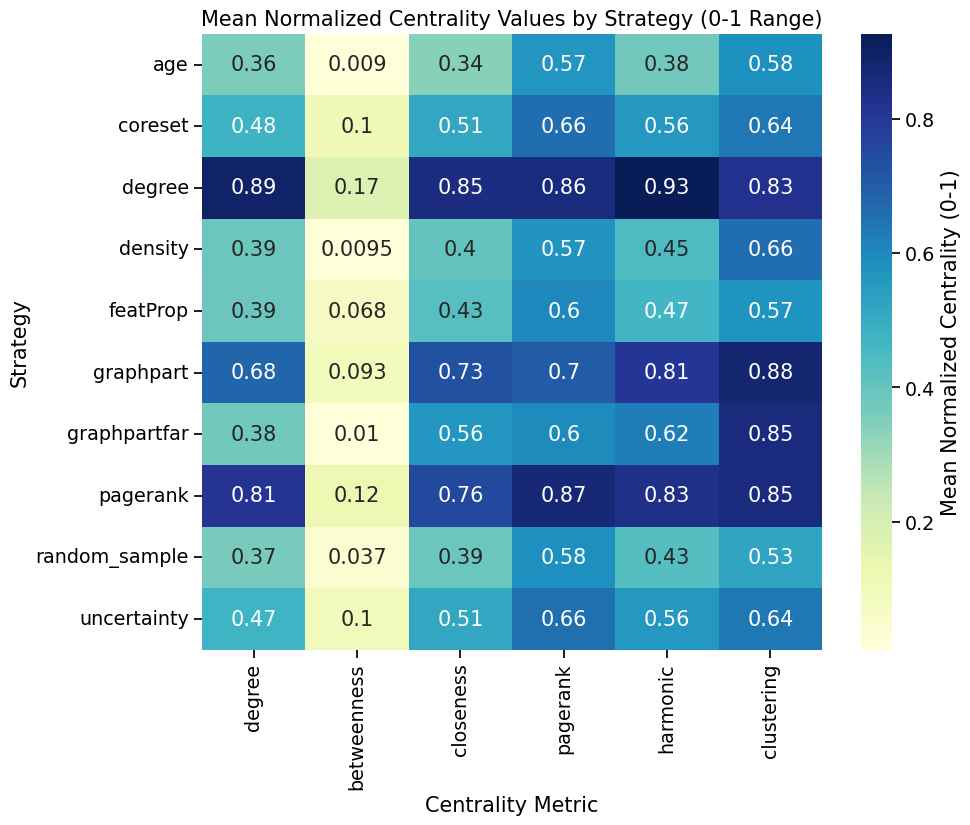}
    \caption{Mean normalized centrality values across different strategies for sleep classification}
    \label{fig:centrality_sleep}
\end{figure}

The heatmap in Figure \ref{fig:centrality_sleep}  visualizes the mean normalized centrality values across various Active Learning strategies for five centrality metrics: degree, betweenness closeness, pagerank, harmonic, and clustering coefficient. The normalization allows for a comparative view of how each strategy targets nodes with specific centrality characteristics.

The degree strategy, as expected, shows high values for degree, closeness, and harmonic centralities, suggesting a strong emphasis on nodes with higher connectivity and central positioning within the network. Similarly, the pagerank strategy exhibits elevated pagerank values, prioritizing influential nodes with significant network connections. This targeted approach indicates that these strategies may focus on well-connected nodes, which could be valuable for tracking widely influencing behaviors but may overlook less central nodes.

On the other hand, strategies like graphpart and graphpartfar demonstrate higher clustering coefficient values, indicating a preference for sampling nodes within tightly-knit groups. This focus on clustered nodes may be useful in capturing localized patterns in sleep behavior, which can be influenced by social connections in closely bonded sub-networks.

Strategies such as age, random sample, and uncertainty appear to take a more balanced or random approach, showing lower values across multiple centrality metrics. This suggests that these strategies may distribute sampling more evenly across the network, which could provide a broader representation of sleep behaviors at the cost of potentially missing insights from highly central or influential nodes. 

\section{Results for SNAPSHOT Dataset}
\subsection{Average Performance Comparison Across Node Types}
The results for sleep classification on the SNAPSHOT dataset and SMS graph are presented in this section, offering a detailed analysis of various Active Learning (AL) strategies and their performance across multiple metrics and node types. Since, the label is relatively balanced, only accuracy and CPI computed for accuracy are reported. Each cohort is treated as a separate network since there was minimal overlap between cohorts. Results for the second cohort are reported below.

\begin{table}[]
\centering
\resizebox{\columnwidth}{!}{%
\begin{tabular}{|p{2.8cm}|p{2.5cm}|p{2.5cm}|p{2.5cm}|p{2.5cm}|}
\hline
\textbf{AL Strategy} & \textbf{Test Set (Same Day)} & \textbf{Unqueried Nodes (Same Day)} & \textbf{Unqueried Nodes (Next Day)} & \textbf{Train Nodes (Next Day)} \\ \hline
random\_sample & 0.72 (0.03) & 0.70 (0.03) & 0.74 (0.03) & 0.75 (0.03) \\ \hline
no AL (baseline) & 0.71 (0.03) & 0.67 (0.04) & 0.65 (0.05) & N/A \\ \hline
graphpart & 0.75 (0.02) & 0.71 (0.03) & 0.70 (0.03) & 0.73 (0.04) \\ \hline
graphpartfar & \textbf{0.77 (0.01)} & 0.73 (0.02) & 0.72 (0.03) & 0.77 (0.02) \\ \hline
age & 0.76 (0.02) & 0.75 (0.01) & 0.76 (0.02) & 0.78 (0.02) \\ \hline
coreset & 0.74 (0.03) & \textbf{0.76 (0.02)} & 0.74 (0.02) & 0.78 (0.03) \\ \hline
featProp & 0.75 (0.02) & 0.75 (0.02) & 0.74 (0.03) & \textbf{0.81 (0.02)} \\ \hline
density & \textbf{0.77 (0.01)} & 0.75 (0.02) & \textbf{0.78} (0.02) & 0.78 (0.02) \\ \hline
uncertainty entropy & 0.75 (0.02) & 0.75 (0.02) & 0.76 (0.03) & 0.76 (0.03) \\ \hline
uncertainty least confidence & 0.75 (0.03) & 0.74 (0.02) & 0.76 (0.03) & 0.76 (0.03) \\ \hline
uncertainty margin & 0.74 (0.04) & 0.70 (0.05) & 0.75 (0.04) & 0.75 (0.03) \\ \hline
pagerank & 0.74 (0.02) & 0.74 (0.02) & 0.74 (0.02) & 0.76 (0.02) \\ \hline
degree & 0.74 (0.02) & 0.75 (0.02) & 0.74 (0.03) & 0.72 (0.02) \\ \hline
\end{tabular}%
}
\caption{CPI Comparison for Active Learning Strategies (Sleep Label, L=8, k=9)}
\label{tab:cpi_comparison_sleep_label_SS}
\end{table}

\begin{table}[]
\centering
\resizebox{\columnwidth}{!}{%
\begin{tabular}{|p{2.8cm}|p{2.5cm}|p{2.5cm}|p{2.5cm}|p{2.5cm}|}
\hline
\textbf{AL Strategy} & \textbf{Test Set (Same Day)} & \textbf{Unqueried Nodes (Same Day)} & \textbf{Unqueried Nodes (Next Day)} & \textbf{Train Nodes (Next Day)} \\ \hline
random\_sample & 0.77 (0.17) & 0.78 (0.13) & 0.80 (0.15) & 0.80 (0.16) \\ \hline
no AL (baseline) & 0.75 (0.18) & 0.72 (0.09) & 0.70 (0.14) & N/A \\ \hline
graphpart & 0.79 (0.17) & 0.75 (0.11) & 0.75 (0.15) & 0.78 (0.18) \\ \hline
graphpartfar & 0.81 (0.16) & 0.78 (0.11) & 0.77 (0.14) & 0.82 (0.16) \\ \hline
age & 0.81 (0.16) & 0.79 (0.11) & \textbf{0.81 (0.12)} & 0.84 (0.14) \\ \hline
coreset & 0.78 (0.16) & \textbf{0.81 (0.13)} & 0.79 (0.14) & 0.82 (0.17) \\ \hline
featProp & 0.79 (0.16) & 0.80 (0.11) & 0.79 (0.13) & \textbf{0.86 (0.14)} \\ \hline
density & \textbf{0.82 (0.16)} & 0.80 (0.10) & \textbf{0.81 (0.11)} & 0.83 (0.14) \\ \hline
uncertainty entropy & 0.79 (0.16) & 0.75 (0.12) & 0.79 (0.13) & 0.78 (0.16) \\ \hline
uncertainty least confidence & 0.79 (0.17) & 0.79 (0.12) & 0.80 (0.14) & 0.81 (0.16) \\ \hline
uncertainty margin & 0.78 (0.18) & 0.75 (0.14) & 0.80 (0.15) & 0.80 (0.16) \\ \hline
pagerank & 0.79 (0.17) & 0.78 (0.12) & 0.79 (0.14) & 0.81 (0.16) \\ \hline
degree & 0.78 (0.17) & 0.80 (0.12) & 0.79 (0.15) & 0.78 (0.15) \\ \hline
\end{tabular}%
}
\caption{Accuracy Comparison for Active Learning Strategies (Sleep Label, L=8, k=9)}
\label{tab:accuracy_comparison_sleep_SS}
\end{table}

The results presented in the CPI (\ref{tab:cpi_comparison_sleep_label_SS}) and Accuracy (\ref{tab:accuracy_comparison_sleep_SS}) tables highlight the performance differences across active learning strategies. Graph-based strategies such as graphpart, graphpartfar, and density consistently outperform other approaches on the test set (same day), demonstrating their ability to generalize effectively to unseen data. For example, in the Accuracy table (\ref{tab:accuracy_comparison_sleep_SS}), graphpartfar achieves a test set accuracy of 0.81 (0.16), while density closely follows with 0.82 (0.16), showcasing their robustness in handling the complexities of graph-structured datasets.

For unqueried nodes, uncertainty-based strategies like entropy and least-confidence show competitive performance, as they prioritize selecting the most uncertain nodes for labeling, ensuring better representation of challenging data points.  Regarding future predictability, strategies such as density, age, and featprop excel for both unqueried and train nodes (next day), with featprop achieving a high train node accuracy of 0.86. This reflects their ability to not only adapt to the present but also provide reliable predictions for subsequent time periods, emphasizing their utility in dynamic, real-world scenarios.

\subsection{Daily Variation Analysis of Active Learning Strategies by Node
Types and Metrics}

\begin{figure}[htbp!]
    \centering
    \includegraphics[width=\textwidth,height=\textheight]{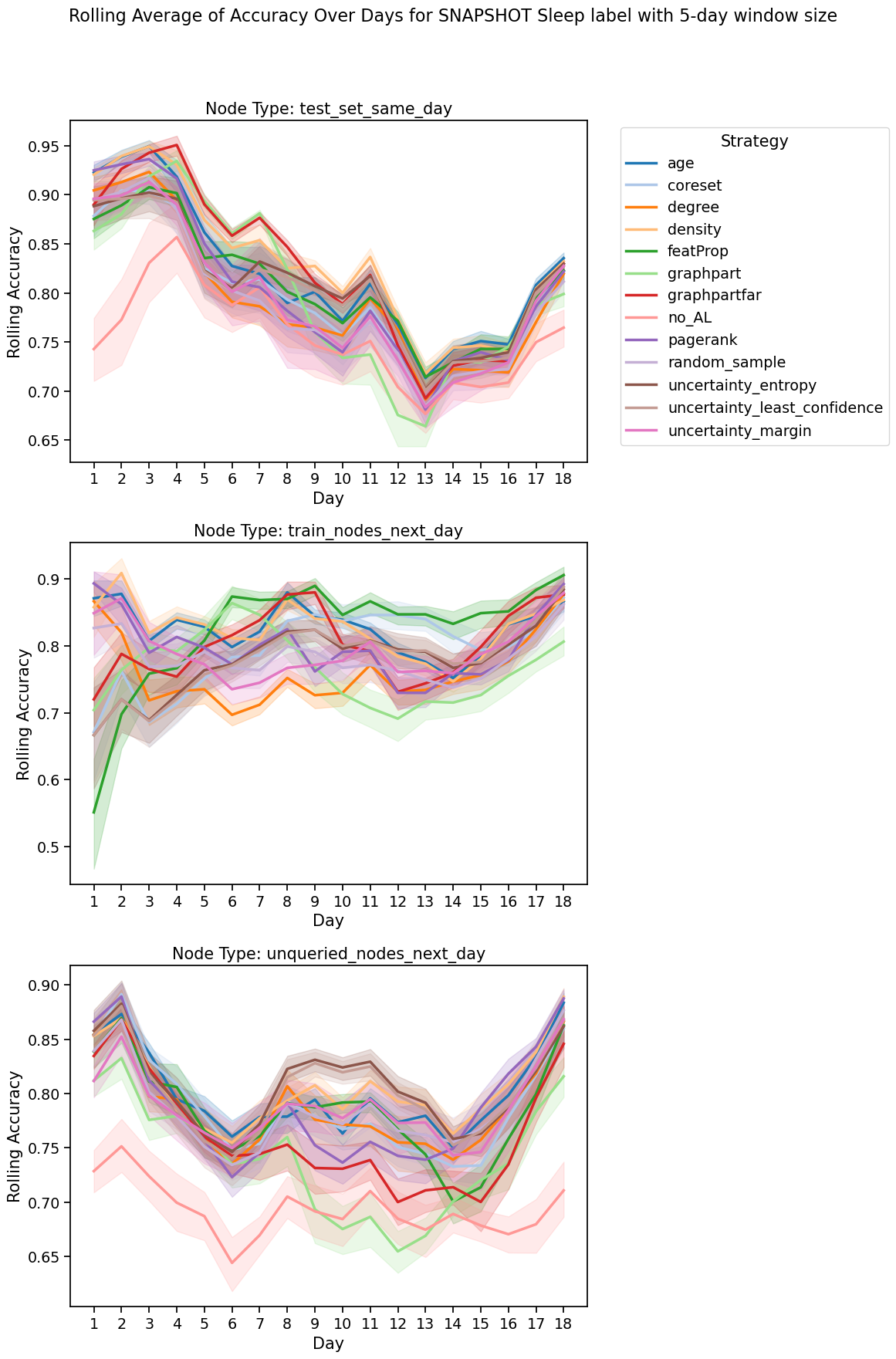}
    \caption{Rolling Average and standard deviation of Accuracy Over Days for SNAPSHOT Sleep Label with 5-day Window Size for L=8 and k=9}
    \label{fig:rolling_sleep_SS}
\end{figure}

The rolling average of accuracy across days in Figure~\ref{fig:rolling_sleep_SS} provides insights into the performance of active learning strategies for different node types: test set (unseen users), train nodes next day (future behavior prediction), and unqueried nodes (not queried in the active learning process). Initially, accuracy begins at a high level, reflecting the model's strong starting point due to its training on all available data in the initial training phase. However, as active learning strategies are implemented and the model starts querying selective nodes, there is a noticeable shift in performance trends. For the test set, accuracy declines over time, reaching its lowest point around the midpoint of the study, and then starts to go up again, likely due to challenges in generalizing to unseen users while adapting to selective querying.

For train nodes, the rolling accuracy exhibits steady improvement after an initial dip, showcasing the model's ability to learn effectively from the limited but informative queried samples to predict future behavior. The unqueried nodes category shows lower accuracy trends compared to the test set, reflecting its lack of direct influence from the training process. Across all node types, periodic fluctuations in accuracy suggest shifts in behavioral patterns or changes in the underlying data, temporarily disrupting model performance before eventual recovery.

The differences in strategy performance become apparent over time, with approaches like graphpartfar, density, and age showing faster recovery and higher overall accuracy, indicating their effectiveness in identifying valuable samples. In contrast, strategies like random sample demonstrate slower improvements, highlighting less efficiency in selecting informative data. As time progresses, most strategies converge toward better performance, underscoring the importance of targeted approaches in maintaining accuracy amidst evolving data patterns. This analysis highlights how active learning strategies influence the model's ability to generalize, adapt, and maintain accuracy over time.

\subsection{Evaluating Diversity \& Sampling Burden across Multiple AL Strategies }

\begin{figure}
    \centering
    \includegraphics[width=1.0\linewidth]{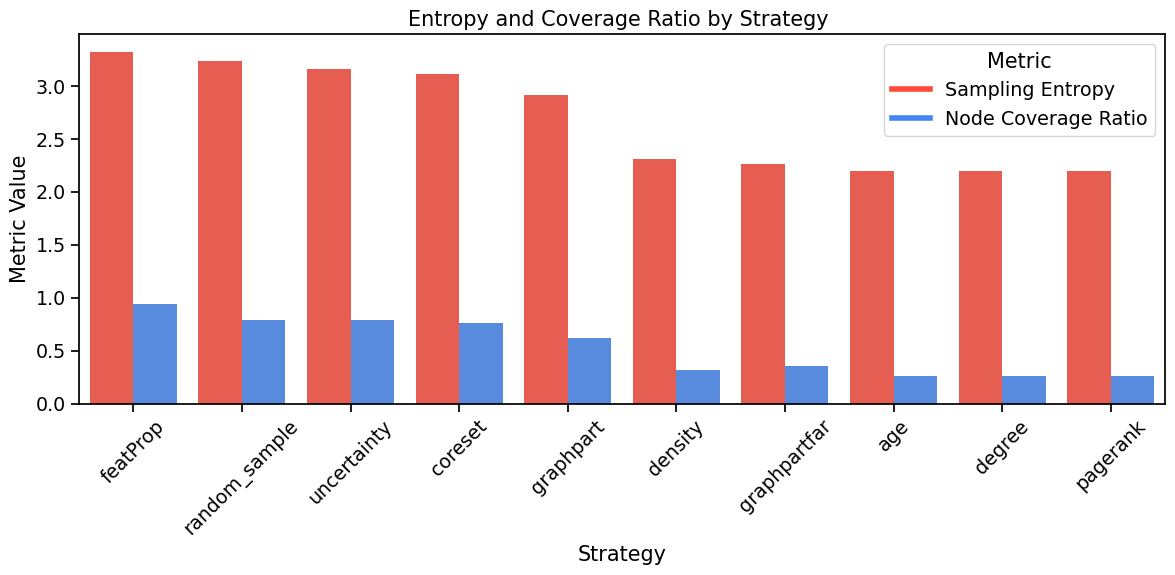}
    \caption{Sampling Entropy and Node Coverage Ratio for various Active Learning Strategies in SNAPSHOT Sleep Classification}
    \label{fig:entropy_coverage_sleep_SS}
\end{figure}

The results in the Figure. \ref{fig:entropy_coverage_sleep_SS} for sleep classification offers insights into the performance of various Active Learning (AL) strategies in terms of Sampling Entropy and Node Coverage Ratio. Strategies like featProp and random sampling maintain the highest Sampling Entropy, indicating that these methods achieve a balanced and diverse selection of nodes across the network. This ensures a representative sampling process. Such diversity is particularly valuable in applications where understanding a wide range of patterns is necessary for accurate and inclusive sleep classification.

In contrast, strategies like PageRank, degree, age, and graphpartfar demonstrate lower Sampling Entropy, reflecting a more focused approach that prioritizes influential or specific subsets of nodes. While this can be advantageous in scenarios where insights from key individuals are critical, it may reduce the diversity of the sampled data. Similarly, Node Coverage Ratio reveals that strategies like featProp and random sampling effectively capture information from a broader range of nodes, ensuring comprehensive network representation. On the other hand, PageRank, degree, age, and graphpartfar have lower coverage, potentially overlooking fewer central nodes and limiting the model's ability to generalize findings across the network.

\begin{figure}
    \centering
    \includegraphics[width=1.0\linewidth]{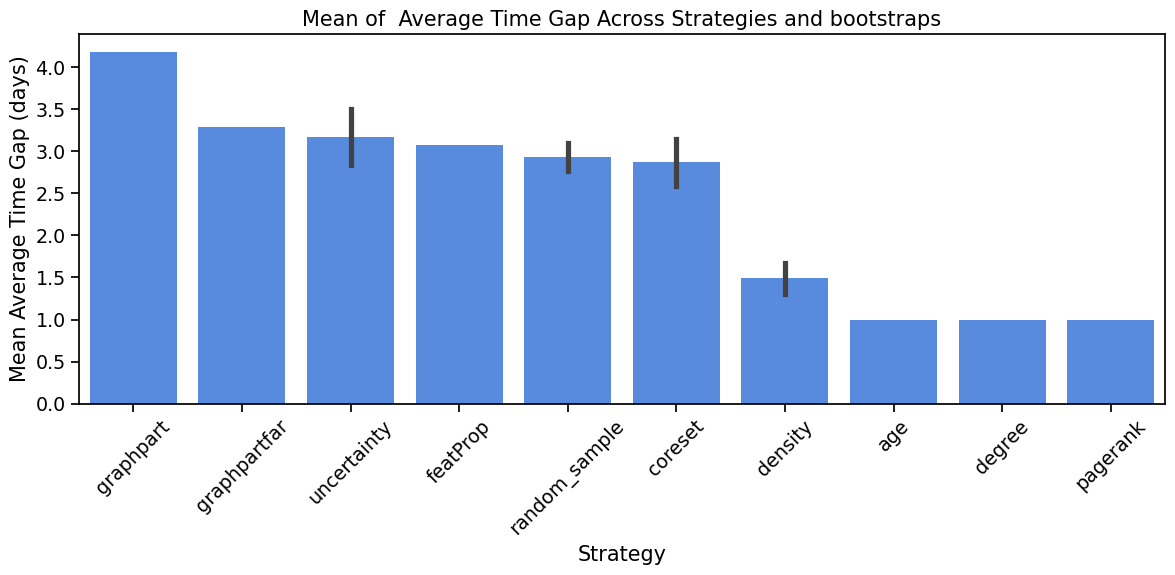}
    \caption{Mean Average Time Gap across Active Learning Strategies for SNAPSHOT Sleep Classification}
    \label{fig:avg_time_gap_sleep_SS}
\end{figure}

The Average Time Gap plot in Figure. \ref{fig:avg_time_gap_sleep_SS} illustrates the mean interval in days between consecutive samples for each node, highlighting how different active learning strategies distribute their sampling temporally. Strategies like graphpart and graphpartfar exhibit the longest average time gaps, indicating that on-average they prioritize spaced-out sampling to reduce the burden on users and allow for the capture of long-term trends. Conversely, strategies such as pagerank, degree, and age show shorter time gaps, reflecting more frequent sampling of specific nodes and a potential focus on capturing immediate changes or influential behaviors.

When compared with the Sampling Entropy and Node Coverage Ratio plot, clear differences in strategy behavior emerge. Strategies like featProp and coreset, which demonstrate high entropy and broad node coverage, maintain moderate time gaps, suggesting a balance between diverse node selection and reasonable sampling intervals. This combination makes them effective for capturing a wide array of data patterns while minimizing user fatigue. On the other hand, strategies such as PageRank and degree characterized by lower entropy and node coverage focus on a smaller subset of nodes with shorter time gaps. While this targeted approach might be effective for tracking key nodes, it risks over-sampling and overlooking broader network dynamics, limiting the representativeness of the overall dataset.

\begin{figure}
    \centering
    \includegraphics[width=1.0\linewidth]{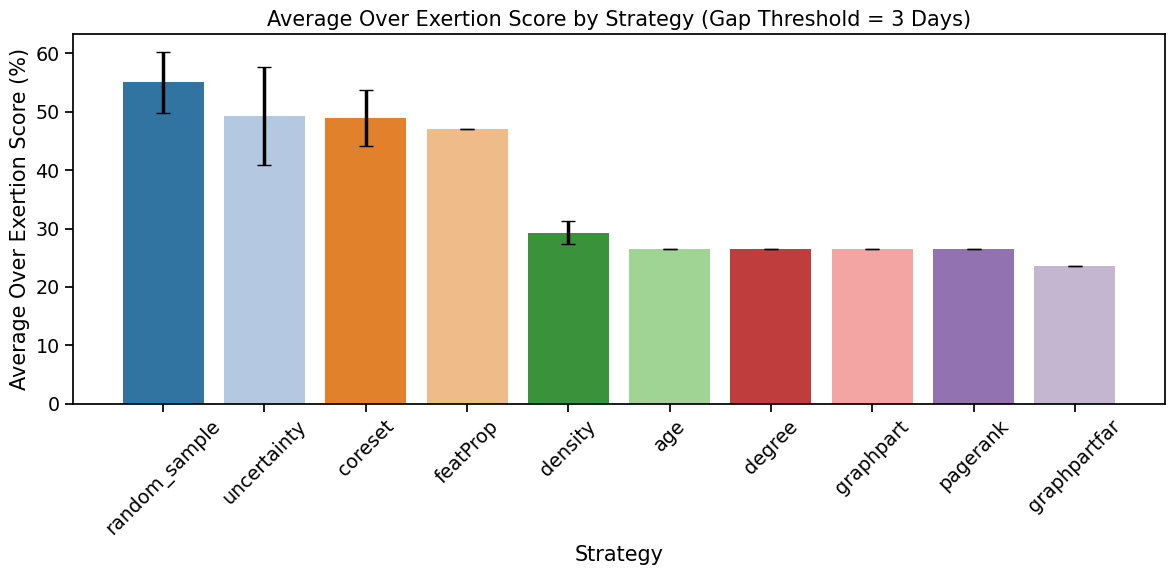}
    \caption{Average Over Exertion Score by strategy at a fixed 3-day gap threshold for SNAPSHOT Sleep Classification}
    \label{fig:over_exer_sleep_SS}
\end{figure}

The average time-gap metric has the potential to overlook the burden caused when a small percentage of users are sampled very frequently, while a larger percentage experiences more spaced-out sampling. This limitation is partially addressed by the exertion score.
 The Average Over Exertion Score for sleep classification (Gap Threshold = 3 Days) in Figure \ref{fig:over_exer_sleep_SS} highlights how frequently each active learning strategy samples nodes within short intervals, potentially leading to user fatigue or resource exhaustion. Strategies such as random sampling, uncertainty, and coreset exhibit higher over-exertion scores, indicating that they often re-sample nodes within the 3-day threshold. This behavior may increase the burden on users or systems, particularly in applications where frequent interaction for query/label is not desirable.

In contrast, strategies like graphpart, pagerank, and graphpartfar demonstrate significantly lower over-exertion scores, reflecting a more restrained and balanced approach to node sampling. These strategies avoid excessive re-sampling, making them better suited for applications that prioritize sustained user engagement and efficient resource allocation over time. 
\begin{figure}
    \centering
    \includegraphics[width=\linewidth]{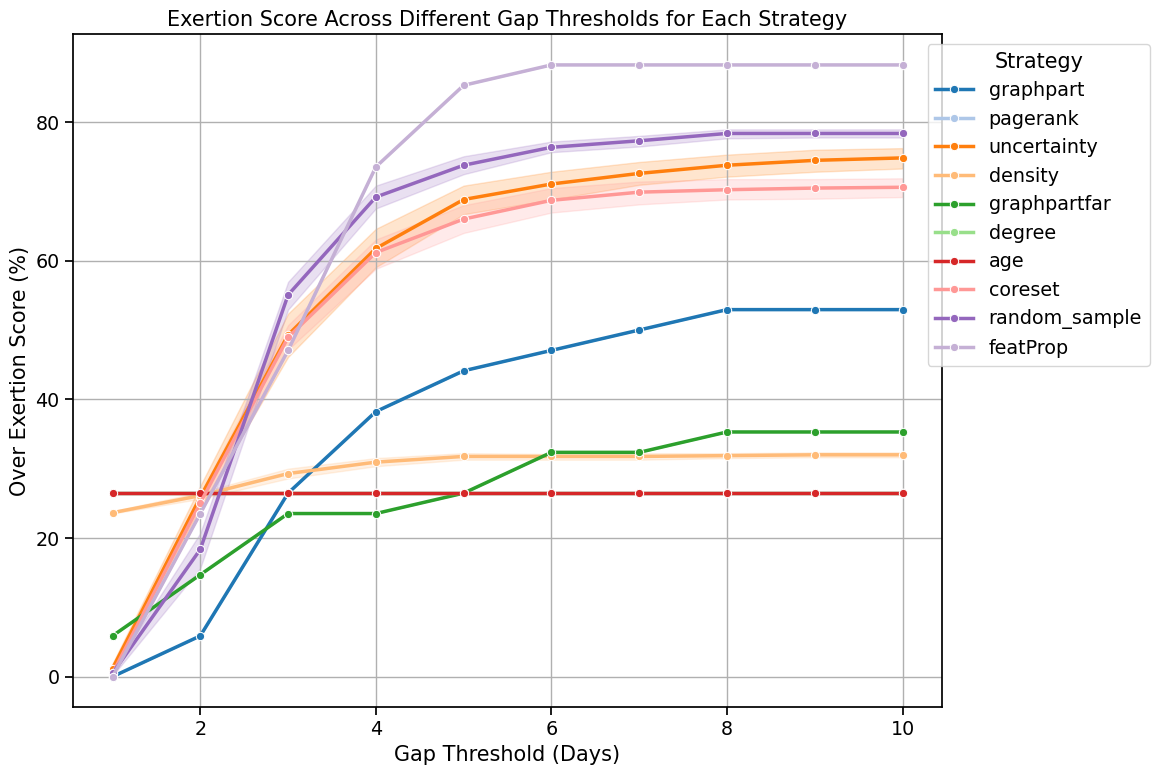}
    \caption{Variation of Over Exertion Score across different gap thresholds for SNAPSHOT Sleep Classification}
    \label{fig:over_exer_sleep_varygap_SS}
\end{figure}

The Over-exertion Score across different gap thresholds in Figure \ref{fig:over_exer_sleep_varygap_SS} further illustrates how each strategy's sampling pattern adapts as the allowable interval between queries increases. While most strategies exhibit rising over-exertion scores with longer gap thresholds, strategies like graphpartfar, age, and density consistently maintain low exertion levels, reflecting their restrained and balanced approach to node sampling. This indicates a focus on evenly distributing queries over time, which can help minimize user fatigue and ensure sustainable data collection.

In contrast, strategies such as featProp, uncertainty, random sampling, and coreset show sharp increases in over-exertion scores as the gap threshold widens, suggesting frequent updates of specific nodes. This behavior may be advantageous for high-resolution monitoring but can impose a heavier burden on the system or users. These trends highlight the need to align the choice of an active learning strategy with the application's objectives, whether emphasizing frequent data collection for detailed tracking or adopting a balanced approach to optimize long-term sustainability.

\subsection{Centrality Preferences and Exertion Patterns
Across Sampling Strategies}

\begin{figure}
    \centering
    \includegraphics[width=1.0\linewidth]{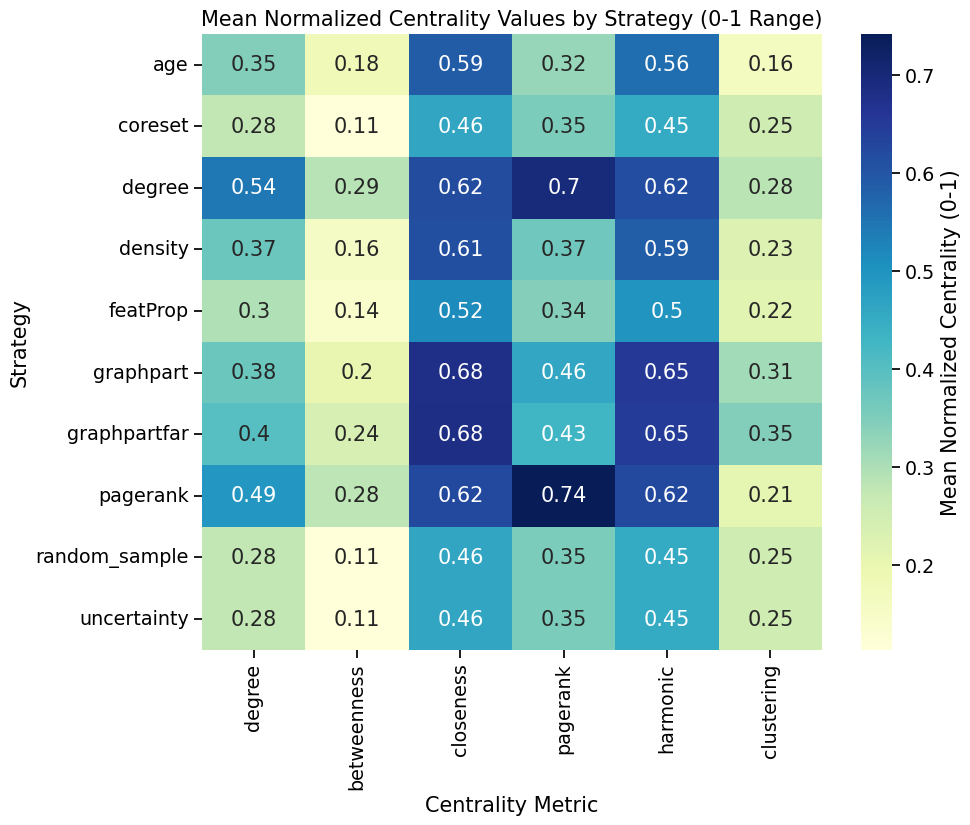}
    \caption{Mean Normalized Centrality values across different strategies for SNAPSHOT Sleep Classification}
    \label{fig:centrality_sleep_SS}
\end{figure}

The heatmap in Figure \ref{fig:centrality_sleep_SS}  visualizes the mean normalized centrality values across various Active Learning strategies for five centrality metrics: degree, betweenness closeness, pagerank, harmonic, and clustering coefficient. The normalization allows for a comparative view of how each strategy targets nodes with specific centrality characteristics.

The degree-based sampling exhibits prominent values for degree, closeness, and harmonic centralities, indicating its focus on highly connected and central nodes in the network. Likewise, the pagerank strategy demonstrates elevated pagerank values, emphasizing influential nodes with significant network connectivity. These strategies are particularly effective for tracking behaviors that propagate through key nodes but might neglect peripheral areas of the network.

In contrast, strategies such as graphpart and graphpartfar show higher clustering coefficient values, revealing a tendency to target nodes within tightly interconnected communities. This preference for localized clusters could be advantageous for uncovering patterns influenced by social connections or interactions within smaller subgroups of the network.

Strategies like age, random sampling, and uncertainty display comparatively lower values across most centrality metrics, indicating a more uniform or dispersed sampling pattern. While this broader coverage ensures diverse representation across the network, it may miss opportunities to capture behaviors originating from highly central or influential nodes.

\section{Statistical Tests for Performance Evaluation}

\begin{table}[h!]
\centering
\begin{tabular}{|p{2.0cm}|p{1.9cm}|p{1.8cm}|p{1.8cm}|p{2.0cm}|p{1.0cm}|p{1.2cm}|}
\hline
\multirow{2}{*}{\textbf{Node Type}} & \multicolumn{2}{c|}{\textbf{Accuracy}} & \multicolumn{2}{c|}{\textbf{F1 Macro}} & \multicolumn{2}{c|}{\textbf{CPI}} \\
\cline{2-7}
& \textbf{ANOVA p-value} & \textbf{Kruskal-Wallis p-value} & \textbf{ANOVA p-value} & \textbf{Kruskal-Wallis p-value} & \textbf{ANOVA p-value} & \textbf{Kruskal-Wallis p-value} \\
\hline
Test set same day         & \(2.92 \times 10^{-232}\) & \(1.33 \times 10^{-205}\) & \(1.11 \times 10^{-168}\) & \(1.40 \times 10^{-174}\) & \(0.0\) & \(0.0\) \\
\hline
Train nodes next day      & \(1.98 \times 10^{-153}\) & \(7.42 \times 10^{-172}\)  & \(6.41 \times 10^{-167}\)  & \(8.38 \times 10^{-145}\)  & \(0.0\) & \(0.0\) \\
\hline
Unqueried nodes next day  & \(6.49 \times 10^{-61}\)  & \(2.06 \times 10^{-51}\)  & \(1.54 \times 10^{-11}\)  & \(1.26 \times 10^{-11}\)  & \(0.0\) & \(1.21 \times 10^{-255}\) \\
\hline
Unqueried nodes same day  & \(5.27 \times 10^{-62}\) & \(2.65 \times 10^{-57}\) & \(8.23 \times 10^{-17}\) & \(2.83 \times 10^{-20}\) & \(0.0\) & \(0.0\) \\
\hline
\end{tabular}
\caption{Statistical Test Results for Node Types by Metric for Multiple AL Strategy Comparison for Sleep Label}
\label{tab:stat_results_node_types_Sleep}
\end{table}

% \bibliography{references}
% \bibliographystyle{icml2021}